%% file: jair-cpmpy.tex
\documentclass[manuscript, screen]{jair}

\setcopyright{cc}
\JAIRAE{TBD}
\JAIRTrack{}
\acmMonth{08}
\acmYear{2026}
\acmDOI{TBD}

\RequirePackage[
  datamodel=acmdatamodel,
  style=acmauthoryear,
  backend=biber,
  giveninits=true,
  uniquename=init
  ]{biblatex}

\usepackage{xspace}%

\usepackage{graphicx}%
\usepackage{multirow}%
\usepackage{amsmath,amsfonts}%
\usepackage{amsthm}%
\usepackage{mathrsfs}%
\usepackage[title]{appendix}%
\usepackage{xcolor}%
\usepackage{textcomp}%
\usepackage{manyfoot}%
\usepackage{booktabs}%
\usepackage{algorithm}%
\usepackage{algorithmicx}%
\usepackage[noend]{algpseudocode}%
\usepackage{listings}%
\usepackage{stmaryrd}
\usepackage{adjustbox}
\usepackage{subcaption}
\usepackage{mathtools}
\usepackage{pgfplots}
\usepackage{enumitem}
\usepackage{makecell}
\usepackage{placeins}
\usepackage{cleveref}

\usepackage{thmtools}
\declaretheorem[name=Definition]{definition}
\crefname{definition}{Definition}{Definitions}

\crefname{Grammar}{grammar}{grammars}
\usepackage{syntax}
\newfloat{Grammar}{htbp}{lop}

\input{auxiliary}

\input{images/waterfall}

\begin{document}

\title{Translating finite-domain integer constraint models to CP/SMT/ILP/PB/SAT solvers with CPMpy}

\author{Tias Guns}
\orcid{0000-0002-2156-2155}
\email{tias.guns@kuleuven.be}
\author{Ignace Bleukx}
\orcid{0000-0001-7810-8351}
\email{ignace.bleukx@kuleuven.be}
\affiliation{%
  \institution{KU Leuven}
  \country{Belgium}
}
\author{Hendrik Bierlee}
\orcid{0000-0001-6766-5435}
\email{bierlee.henk@gmail.com}
\affiliation{%
  \institution{KU Leuven}
  \country{Belgium}
}

\author{Jo Devriendt}
\orcid{0000-0002-6346-3665}
\email{jo.devriendt@nonfictionsoftware.com}
\affiliation{%
  \institution{Nonfiction Software}
  \country{Belgium}
}

\author{Emilio Gamba}
\orcid{0000-0003-1720-9428}
\email{emilio.gamba@flandersmake.be}
\affiliation{%
  \institution{Flanders Make}
  \country{Belgium}
}

\author{Orestis Lomis}
\orcid{0009-0006-2934-0510}
\email{orestis.lomis@kuleuven.be}
\affiliation{%
  \institution{KU Leuven}
  \country{Belgium}
}

\author{Wout Piessens}
\orcid{0009-0005-5608-2097}
\email{wout.piessens@kuleuven.be}
\affiliation{%
  \institution{KU Leuven}
  \country{Belgium}
}

\author{Thomas Sergeys}
\orcid{0009-0003-3817-4516}
\email{thomas.sergeys@kuleuven.be}
\affiliation{%
  \institution{KU Leuven}
  \country{Belgium}
}

\author{Dimos Tsouros}
\orcid{0000-0002-3040-0959}
\email{dtsouros@uowm.gr}
\affiliation{%
  \institution{University of Western Macedonia}
  \country{Greece}
}

\author{Wout Vanroose}
\orcid{0009-0004-8945-0442}
\email{wout@vanroose.be}
\affiliation{%
  \institution{KU Leuven}
  \country{Belgium}
}

\author{Hélène Verhaeghe}
\orcid{0000-0003-0233-4656}
\email{helene.verhaeghe@uclouvain.be}
\affiliation{%
  \institution{UCLouvain}
  \country{Belgium}
}

\renewcommand{\shortauthors}{Guns and Bleukx et al.}

\begin{abstract}
{\bf Background:} 
Constraint solving is a declarative approach for solving combinatorial satisfaction and optimization problems.
The user specifies their problem through constraints and decision variables, and a generic solver is used to find a solution.
Several constraint-solving technologies exist, and certain solvers perform well on certain problems.
Therefore, it is useful to try different solvers given a particular application.
However, each solving paradigm supports different types of constraints and decision variables.

{\bf Objectives:}
Our goal is to translate high-level constraint satisfaction and optimization problems into any lower-level formalism, including CP, SMT QF-LIA, ILP, PB and (Max)SAT.
This allows for comparing different solving technologies for a particular problem, without requiring a user to manually remodel it for each solving paradigm.

{\bf Methods:}
We define a high-level language of logical and arithmetic operations, and useful additional functions and constraints, which are known as \textit{global constraints} in the CP community. 
We then present a modular framework for transforming our high-level modeling language to CP/SMT/ILP/PB and (Max)SAT solvers.
While many transformations are partly described in the literature, we observe that they can be implemented through a modular \textit{waterfall} of smaller components, where lower-level paradigms reuse the transformations of higher-level paradigms. 
Two recurring challenges are handling the negation of arbitrary subexpressions and avoiding the introduction of auxiliary variables.
Additionally, we take special care linearizing non-linear operators for ILP, PB and SAT-solvers.

{\bf Results:}
The transformation waterfall is implemented and evaluated in the open-source CPMpy library.
Our results show that constraint models significantly change throughout the transformations, and that optimizations to the linearization of constraints are essential for ILP and PB solvers.
    
\end{abstract}

\received{07/08/2026}

\keywords{modeling languages, constraint reformulation, constraint solving, constraint satisfaction}

\maketitle

\section{Introduction}

A prevalent AI approach to solving combinatorial satisfaction and optimization problems is the \textit{model + solve} paradigm. 
In this paradigm, the user first describes their problem in a constraint model in terms of \textit{constraints} over \textit{decision variables}, optionally with an objective function over a subset of the decision variables. 
This declarative problem specification is then given to a highly optimized \emph{constraint solver}, which searches for an (optimal) solution, where the constraints are used to effectively reduce the search space.

Many generic solvers have been developed, differing in the type of constraints and decision variables they support. 
For example, SAT solvers support Boolean variables and \textit{clauses} as constraints~\cite{biere2021handbook}. 
MaxSAT solvers additionally support a weighted linear sum as an objective function~\cite{li2021maxsat}. 
Pseudo-Boolean solvers support linear constraints over Boolean variables~\cite{roussel2021pseudoboolean}.
There are also Integer Linear Programming (ILP) solvers, Mixed continuous-Integer Linear Programming (MILP) and non-linear programming solvers, Constraint Programming (CP)~\cite{rossi2006handbook} solvers with global constraints, and SAT Modulo Theories (SMT) solvers with arbitrary theories including quantified formulas and unbounded variables~\cite{barrett2021smt}.

We here focus on discrete, \textit{finite-domain} integer constraint models, which have a wide range of real-life applications, including scheduling, packing, assignment, allocation, routing, and more~\cite{schwerin1997binpacking,clautiaux2008orthogonal,fromherz2001scheduling,vanbeek1999cplan,debacker2000vrp}. 
The supported decision variables are Boolean and finite-domain integer. 

Constraint solvers are typically categorized by the types of \textit{constraints} they support, that is, relations they can enforce to be satisfied (such as $a \lor b$, $2x + 3y \leq 4$, $x \neq y$). 
In contrast, SMT solvers and modeling languages are defined by the \textit{expressions} they support, where basic expressions can be combined into nested expressions, which together with a comparison or Boolean operator become a constraint that must be satisfied (such as $a \implies(\neg(b \land c) \lor d)$ or $x\cdot y + 3\cdot(z = 0) \leq \frac{r}{10}$).
What we will describe is a modeling language, with a practical implementation that is embedded in the Python programming language, with the following types of expressions:
\begin{itemize}
    \item Boolean logical operators \textsc{(and, or, not, implication, equivalence, exclusive-or)}
    \item Integer arithmetic operators \textsc{(+, -, multiplication, integer-division, modulo, power, max, min, abs)}
    \item Integer comparison \textit{($=, \neq, >, \geq, <, \leq$)}
    \item Integer ``global'' constraints \textsc{(AllDifferent, Table, Circuit, NoOverlap, Cumulative, LexLess ...)}
    \item Integer ``global'' functions \textsc{(Array-Lookup, Count, Among, NValue ...)}
\end{itemize}
The ``global'' constraints are constructs from the constraint programming community~\cite{globalconstraintcatalog} that offer two benefits: 
1) they provide meaningful abstractions that users can use to compactly write down their problem specification; 
2) many CP solvers have specialized reasoning methods (propagators) to efficiently reduce the search space based on the semantics of the global constraint.
For solvers that do not support such global constraints, there are well-known translations into basic Boolean and integer operators (called decompositions). 
The integer global constraints are Boolean-valued (they can be true or false). 
The ``global'' functions are integer-valued functions that have a corresponding global constraint, e.g., the $\cons{Count}{\mathit{Arr}, \mathit{Val}}$ operator has $\cons{CountEquals}{\mathit{Arr}, \mathit{Val}, \mathit{Cnt}}$. 
Our choice of language follows from the combinatorial scheduling, routing, and assignment type of applications we target.

A number of modeling languages and transformation libraries exist for combinatorial problems, including MiniZinc~\cite{nethercote2007minizinc}, Essence Prime~\cite{akgun2022conjure} in combination with Savile Row~\cite{nightingale2022savilerow}, PicatSAT~\cite{zhou2015picat}, and Pyomo~\cite{hart2011pyomo}.
They differ in 1) the type of expressions they support, and 2) the type of solvers they can translate to.

\newcommand{\yes}{\checkmark}
\newcommand{\no}{-}
\begin{table}[h]
    \centering
    \caption{Comparison of different constraint modeling systems and their capabilities.}
    \begin{tabular}{l|ccccc}
                      & Solvers & Language & \makecell{Global constraints} & \makecell{Global  functions} & \ignore{$\forall, \exists$ &}  \makecell{Variables} \\ \hline
        Savile Row    & CP/SMT/SAT & Text-based  & \yes & \yes & \ignore{\yes &} Bool/int \\
        MiniZinc      & CP/ILP/FZN-based & Text-based  & \yes & \yes & \ignore{\yes &} Bool/int/float \\
        PyCSP3        & CP  & Python & \yes & \yes & \ignore{\no  &} Bool/int \\
        PicatSAT      & SAT & Text-based  & \yes & \no & \ignore{\no &} Bool/int \\
        Pyomo         & ILP/SMT & Python & \no  & \no  & \ignore{\no &} Bool/int/float \\
        OR-Tools API  & CP  & Python & \yes & \yes & \ignore{\no &} Bool/int \\
        Gurobi API    & ILP  & Python & \no  & \no  & \ignore{\no &} Bool/int/float \\[0.2em]
        CPMpy (ours) & CP/SMT/ILP/PB/SAT & Python & \yes & \yes & \ignore{\no &} Bool/int
    \end{tabular}
    \label{tab:capabilities}
\end{table}

The system closest to CPMpy that we will describe here is the Savile Row system for the Essence Prime language. 
The Savile Row system translates constraint models in the Essence language to CP solvers, SAT solvers and SMT solvers.
However, Essence Prime is a text-based modeling language, whereas our CPMpy modeling language is embedded in Python.
Another important difference is that Savile Row has different backends for the 3 types of solvers, where they reuse low-level components up to some extent, which corresponds to how it has been developed over time.
In contrast, we describe a holistic, hierarchical approach that aims to maximize the number of transformations that can be reused for different solver paradigms and solvers. 
Another highly related technology is PyCSP3, which, like CPMpy, is also a Python-embedded system, but it only translates to the XML-based CP language XCSP3, not to other solver paradigms. 
The Pyomo system is also relevant to our framework.
While it is embedded in Python and translates to multiple MIP, ILP, and SMT solvers, the system does not compile ``down'' to PB or SAT solvers.
Additionally, global constraints and operators are not supported by the Pyomo framework as they are specific to CP modeling.
Finally, the well-known MiniZinc system~\cite{nethercote2007minizinc} is highly related as well. 
Together with Essence Prime, it has been an inspiration both conceptually and academically, based on the different papers and reports published on these systems. 
MiniZinc is a text-based language that also supports floating-point decision variables and floating-point arithmetic. 
Its main output language is the text-based FlatZinc constraint language, which supports (custom definable) flat (= non-nested) constraints over floating-point, integer, and Boolean decision variables. 
The key difference is that the MiniZinc compiler does not translate directly to the language of a specific solver.
Rather, it translates to a flat intermediate language, and this intermediate language must then be reformulated to a specific solver in a separate program (e.g., PicatSAT~\cite{zhou2015picat}, fzn2omt~\cite{contaldo2020minizinc2omt}...).
The latter requires rebuilding certain data structures and modeling patterns, and potentially undoing some of the flattening done by the MiniZinc compiler. 
The translation to integer linear and mixed integer programming solvers can be handled by MiniZinc through configuration options that change how different constraints are decomposed/rewritten \cite{nethercote2007minizinc,belov2016improved}.

We propose a more holistic approach, and adopt the view that solvers (or solver APIs, more specifically) accept a subset of our language as input, and our goal is to \textit{minimally} rewrite our expressions so that only supported expressions remain. 
Furthermore, multiple families of solvers require the same rewrite operations: only CP solvers support global constraints, both ILP and PB solvers benefit from good linearizations, and both PB and SAT solvers require the conversion of integer decision variables to Boolean decision variables.
As we will describe below, the result is a hierarchical \textit{waterfall} of transformations that can effectively rewrite models in the above language to any of the mentioned solvers.

Technically, some solver families will benefit from different rewrite choices than others, most notably integer linear programming solvers, which benefit from global constraint decompositions that lead to strong LP relaxations, for example, that reason over a direct Boolean encoding of the categorical integers. 
The proposed approach is modular and allows for such custom decompositions as we describe in \Cref{sec:linearize}. 
In general, the performance of lower-level solvers can be greatly impacted by the choice of rewrites and the encodings used~\cite{vielma2015mipformulations,tamura2009compiling,walsh2000sat}. 
While the transformations we describe are based on techniques described and validated in the literature, we make no attempt to claim they are the optimal choice. 
Indeed, determining the optimal choice of encodings is an ongoing active research topic in the specific solver fields of research~\cite{akgun2022conjure}.
Instead, we believe that constraint-solving communities can benefit from having accessible libraries that offer good \textit{baseline} encodings. 
As the knowledge of encoding choices further develops, modeling system developers can take over the best practices and make them more widely available.
Furthermore, the modular nature of the proposed approach makes modeling systems a good framework for investigating the effect of different encodings.

Our key contribution is hence not any specific transformation, but the presentation of a modular end-to-end translation pipeline that covers five different constraint-solving paradigms and that has been validated in real applications~\cite{bleukx2025modeling,foschini2026hmlv} and competitions~\cite{audemard2024xcsp3,audemard2025xcsp3}.

\section{Background}

A Constraint Satisfaction Problem (CSP) is a triple $(\variables, \domains, \constraints)$~\cite{rossi2006handbook}.
Here, $\variables$ is a set of decision variables and $\domains$ is a set of finite and discrete domains, which detail the allowed values for each variable.
For any integer variable $x$, $\domainof{x} \subset \mathbb{Z}$ and Boolean variables have a domain $\set{\mathit{true}, \mathit{false}}$.
In our constraint modeling language, only \emph{contiguous domains} are allowed, and each variable $x$ has a domain $\range{\lb{x}}{\ub{x}}$ where $\lb{x}$ and $\ub{x}$ are the lower and upper bounds of $x$, respectively.
An assignment maps a variable to a value in its domain, which we write as $\assign{x}{v}$.
$\constraints$ is a set of constraints, where each constraint $c \in \constraints$ maps an assignment to \emph{true} or \emph{false}.
A constraint is typically written as a mathematical expression or predicate and often only contains a subset of the variables in the full CSP.
E.g., the constraint $x + 2y -8z \geq 3$ only contains variables $x,y$ and $z$.
These variables are said to be in the \emph{scope} of the constraint.
To refer to the set of variables occurring in $c$, we write $\scope{c}$.
If an assignment is mapped to \emph{true} by the constraint, we say the assignment \emph{satisfies} the constraint.
An assignment $\alpha$ \emph{projected to} a subset of variables $\variables$ is written as $\assignmentto{\mathcal{\variables}}$.

An assignment that satisfies all constraints in the CSP is a \emph{solution} of the CSP.
If there exists no solution to the CSP, it is said to be \emph{unsatisfiable}.
The set of solutions of a CSP is written as $\sols{\variables, \domains, \constraints}$ or, when the variables and domains are clear from the context, we write $\sols{\constraints}$ in short.
The set of solutions of a set of constraints, projected to a set of variables $\variables$ is written as $\solsto{\variables}{\constraints}$.

A Constraint Optimization Problem (COP) generalizes the definition of a CSP to also include an objective function $f$.
That is, a COP is a quadruple $(\variables, \domains, \constraints, f)$ where $f$ is a numerical expression that maps a full assignment $\alpha$ to a numerical value.
This allows us to \emph{score} or \emph{rank} the set of solutions to a constraint problem and allows searching for an \emph{optimal} solution of the problem.
An optimal solution (w.l.o.g.) \emph{minimizes} the objective value.
That is, a solution $\alpha^*$ is optimal if for each $\alpha \in \sols{\constraints}$, $f(\alpha)$ is greater than or equal to $f(\alpha^*)$.

\subsection{Constraint reformulations}
A constraint satisfaction or optimization problem (hereafter referred to as ``constraint problem'') can be modeled in multiple ways by choosing a particular set of variables and constraints.
We will be interested in semantically equivalent formulations of a constraint problem.
We say two CSPs are semantically equivalent to a set of decision variables if they have the same set of solutions when projected onto that set of decision variables.
We say two such CSPs are \emph{constraint reformulations} of one another, as formalized by~\citet{vanhentenryck2003be}.

\begin{definition}[Constraint reformulation]\label{def:reformulation}
A CSP $(\variables', \domains', \constraints')$ is a \emph{constraint reformulation} of another CSP $(\variables,\domains,\constraints)$ if and only if both CSPs are semantically equivalent with respect to the set of original decision variables $\variables$.
That is, if $\solsto{\variables}{\variables,\domains,\constraints} = \solsto{\variables}{\variables',\domains',\constraints'}$ and $\variables \subseteq \variables'$.
\end{definition}

An example of a constraint reformulation for $\set{\consmin{x,y,z} \leq 10}$ is $\set{\consmin{x,y,z} = m, m \leq 10}$ where $\variables = \set{x,y,z}$ and $\variables' = \set{x,y,z,m}$.
To write a CSP to a reformulated CSP, we say it has to be \emph{transformed} into the new CSP.
Therefore, we will refer to the process of rewriting a CSP to a constraint reformulation as a \emph{transformation}.
Transformations can introduce new variables into the CSP, such as $m$ in the reformulation of the constraint above, and $\variables' \setminus \variables$ in general.
These fresh variables are introduced by the modeling system and do not belong to the set of user-defined variables.
We call such variables~\emph{auxiliary variables}~\cite{smith2006modelling}.
The values of these auxiliary variables do not need to be uniquely determined by a solution over $\variables$.
In this case, for an original solution there may exist multiple \textit{extensions} to a solution of the reformulated CSP. 
In general, the reformulation must be sound and complete with respect to CSP $(\variables,\domains,\constraints)$: every solution of the reformulated CSP must project onto a solution of the original CSP, and every solution of the original CSP must admit at least one extension to a solution of the reformulated CSP.

A constraint reformulation generalizes to optimization problems as well.
That is, a COP $(\variables', \domains', \constraints', f')$ is a reformulation of $(\variables, \domains, \constraints, f)$ if the set of constraints are semantically equivalent to $\variables$ and for each solution $\alpha \in \sols{\variables',\domains',\constraints'}$, it holds that $f'(\alpha) = f(\assignmentto{\variables})$

Note that in practice, constraint modeling systems can also \textit{remove} variables from the input CSP when transforming it to a CSP suitable for a particular solver.
We distinguish three cases where variables are removed:
\begin{enumerate}
    \item Unconstrained variables, also known as \emph{free} variables, can be removed from the model.
    \item Constant variables with a singleton domain can be replaced by that single value in their domain.
    \item Functionally defined variables whose value is determined by a set of (auxiliary) variables $\variables''$.
\end{enumerate}

Whenever such variables are removed during the transformation process, we assume the modeling system keeps a mapping from the remaining variables to the removed variables.
The modeling system then only posts the \emph{relevant} part of the transformed CSP to the solver (e.g., only Boolean encoding literals of integer variables for SAT solvers), and derives the value for all removed decision variables after the solver has found a solution.

\subsection{Functional constraints}

Some constraints used in constraint models represent a function.
That is, they \emph{define} the value of one or more ``output'' variables, given a value to the ``input'' variables.
Functional constraints can be either \emph{total} or \emph{partial}.

\begin{definition}[Total function constraint] \label{def:total}
A functional constraint $F(\xvar,\yvar)$ is total if and only if for \textit{each} assignment to $\xvar$, there exists \emph{exactly one} assignment to $\yvar$ that satisfies the constraint.
\end{definition}
Total function constraints with a single input and output variable are sometimes also referred to as \emph{views} in constraint programming nomenclature~\cite{vanhentenryck2014views}.
An example of such a total function with a single ``output'' variable is the $\consmin{X,m}$ constraint, as for assignment to the set of variables $X$, the minimum value always exists and is unique.
Another example is the $\gcc{X,V,O}$ constraint, which enforces the number of occurrences of each value $v_i$ in the array of $X$ to be exactly $o_i$.
The \emph{open} version of this constraint allows values in $X$ to take a value not occurring in $V$, i.e. it allows $\sum_{o_j \in O} o_j \leq |X|$~\cite{vanhoeve2006open}.
This constraint is an example of a total function constraint with multiple output variables $O$.
Indeed, for any value assigned to the variables in $X$ and a fixed set of values $C$, the number of occurrences $O_v$ of value $v$ in $X$ is always defined and unique.

\begin{definition}[Partial function constraint] \label{def:partial}
A functional constraint $F(\xvar,\yvar)$ is partial if for \textit{each} assignment to $\xvar$, there exists \emph{at most one} assignment to $\yvar$ that satisfies the constraint.
\end{definition}

This means the output of a partial function may be \emph{undefined} for some input values.
An example of a partial function is the $\textsc{Division}$ function, for which the output is undefined when the denominator is 0.
Note that the \textit{closed} version of $\gcc{X,V,O}$, which enforces values of $X$ to be assigned to a value in $V$, can be interpreted as a partial function.
Indeed, given an assignment to $X$, the constraint only defines the number of occurrences for each value in $V$ when all variables $X$ take a value occurring in $V$.
\Cref{sec:safening} elaborates further on the semantics of (nested) partial functions in constraint models.

\section{Language}
\input{grammars/core-alt}

We now formally define the constraint modeling language as shown in \Cref{gra:cpmpy}, which is used throughout the remainder of this paper.
The modeling language consists of Boolean and integer expressions, which can be arbitrarily nested to form complex expressions.
We say a Boolean expression is a constraint when it is ``top-level'', i.e., if it occurs in the list of Boolean expressions in the model definition.
If a Boolean expression occurs as a subexpression, we say it is ``nested''.
Naturally, integer expressions only occur as a subexpression, as they cannot be enforced to \ttrue at the top-level of the model. 
E.g., $x + y$ is not a constraint.

\subsection{Core language}

We first define the core language of finite-domain integer modeling, that is, everything in Grammar~\ref{gra:cpmpy} except global constraints and functions.
The language allows for two types of decision variables: \emph{Boolean} variables and \emph{integer} variables.
Each integer variable is defined with a name, a lower bound, and an upper bound.
Boolean variables are defined with a name, and the implicit domain $\set{\tfalse, \ttrue}$

Similar to variables, each \emph{expression} has either a Boolean or an integer return type.
Boolean variables (and expressions) can be used in logic operators from propositional logic: negation ($\neg$), conjunction ($\wedge$),  disjunction ($\vee$), implication ($\implies$), and reification/equivalence ($\reifies$).
Integer variables (and expressions) can be used as arguments to the comparisons $\leq, <, \geq, >, =$ and $\neq$.
Operators with an integer return type are the linear arithmetic operators plus ($+$) and minus ($-$) and the multiplication operator \texttimes\xspace to ``scale'' an integer expression with a constant. Other Boolean and integer operators will be implemented through global constraints and global functions.

Note that our language does not support casting from integer to Boolean.
That is, Boolean variables and expressions can be used as an integer (where False and True are interpreted as 0 and 1, respectively), but not vice versa.
E.g., it is not valid to write $\neg(x + y)$, even if the expression $x + y$ is bounded to only take values 0 or 1.

This is similar to the format supported by SMT-solvers~\cite{barrett2010smt,barrett2021smt}.
However, in contrast to the language supported in SMT, our constraint modeling language includes the use of global constraints and global functions, as described in the next section.

\subsection{Global constraints and functions}
We now extend the core language with additional Boolean and integer operators and functions. 
What differentiates them from the core operators is that they can be \emph{decomposed}, i.e., rewritten into (a conjunction of) core operators. 
To stay somewhat close to constraint programming nomenclature, we will call these additional operators and functions ``global constraints'' if they have a Boolean return type, and when they have an integer return type we will call them ``global functions''.

\begin{definition}[Global constraint]
A \emph{global constraint} is a Boolean predicate that can be \emph{reformulated} to a semantically equivalent (see \Cref{def:reformulation}) set of constraints in the core language.
\end{definition}

For example, the global constraint $\talldiff(x_1,\ldots,x_n)$ holds if all variables take pairwise distinct values.
It can be decomposed into the following conjunction of core constraints:
$
\bigwedge_{1 \leq i < j \leq n} x_i \neq x_j.
$
Other examples of well-known global constraints are \tgcc, \tcumulative, \ttable and \txor.
In \Cref{sec:decompose} we elaborate on the semantics of these constraints and their decompositions implemented in our framework.

\begin{definition}[Global function]
A \emph{global function} is an integer-valued function that can be \emph{reformulated} to a semantically equivalent function and set of supporting constraints in the core language.
\end{definition}

Examples of global functions supported in our language are the \tmin, \tabs and \telement functions.
Note that our use of ``global functions'' is often not described in existing papers on constraint modeling languages; they often consider the operator together with an equality operator as a global constraint and silently extract it when the function is used as a subexpression. 
However, treating it as a function is a first key aspect in avoiding unnecessary introductions of auxiliary variables~\cite{stuckey2013minizincwithfunctions}.

\begin{figure}[h]    
\begin{subfigure}[t]{0.30\textwidth}
\centering
\small
\begin{tikzpicture}[
  level distance=0.55cm,
  sibling distance=2cm,
  every node/.style = {shape=rectangle, draw, align=center}
]
\node{\small $\reifies$}
child { node {\small $\vee$}
        [sibling distance=0.8cm]
    child { node {\small $a$}}
    child {node {\small $b$}}
}
child { node {\small $\leq$}
        [sibling distance=0.8cm]
    child { node {\small $+$}
        child {node { \small $x$}}
            child {node {\small $\wedge$}
                child {node {\small $p$}}
                child {node {\small $q$}}
            }
     }
     child {node {\small 3}}
};
\end{tikzpicture}
\caption{$(a \vee b) \reifies (x + (p \wedge q) \leq 3) $}
\end{subfigure}
\begin{subfigure}[t]{0.30\textwidth}
\centering
\small
\begin{tikzpicture}[
  level distance=7mm,
  sibling distance=1.2cm,
  every node/.style = {shape=rectangle, draw, align=center, minimum height=5mm}
]
\node{\small \talldiff}
    child { node {\small $a$}}
    child { node {\small \term{Div}}
        [sibling distance=0.6cm]
      child { node {\small $b$}}
      child { node {\small $c$}}
    }
    child { node {\small +}
        [sibling distance=0.6cm]
      child { node {\small $d$}}
      child { node {\small 3}}
    };
\end{tikzpicture}
\caption{$\alldiff{a,\cons{Div}{b,c}, d + 3}$}
\end{subfigure}
\caption{Examples of constraints in our constraint language and their corresponding expression trees.}
\end{figure}
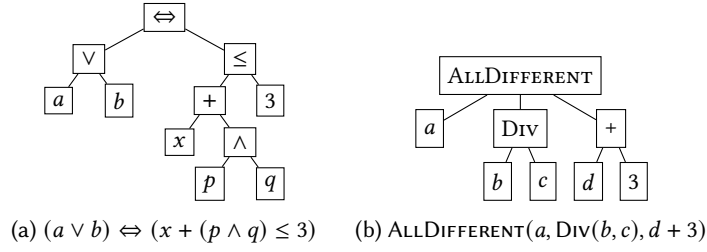

\section{Transformations}\label{sec:transformations}

\transformations{}

We now describe the modular transformation pipeline that allows us to translate the described language to different solver paradigms.
The goal of our transformation process is to transform each constraint in the input constraint model to a set of \emph{primitive constraints} for the targeted solver.
A primitive constraint is a constraint the solver accepts as input.
Naturally, each solver (paradigm) has a certain set of such primitive constraints.

Each stage in the transformation pipeline shown in \Cref{fig:transformations} has a specific purpose of reformulating certain types of constraints.
Each box corresponds to a particular subsection describing the transformation.
A key insight of our transformation pipeline is that lower-level solvers can re-use transformations that are shared with higher-level solvers.
The first step for all solvers is to eliminate partial functions, which is described in \Cref{sec:safening}.
This ensures all subsequent transformations and solver interfaces can assume every global function in the constraint model is a total function.
Next, negation operators are removed from the expression tree as described in \Cref{sec:negation}.
More specifically, we follow a set of standard reformulation rules that allow to ``push down'' any negation operator into its arguments.
Note that this step is omitted for SMT solvers, as they typically have their own set of simplification rules for eliminating negation from the expression tree.

As a third general transformation step, we decompose all unsupported global constraints and functions (\Cref{sec:decompose}).
Here, we replace every unsupported global constraint with a Boolean expression consisting of only supported operators by the solver at hand, and all global functions are replaced with a numerical expression.
After eliminating all these global constraints and functions, we can post the constraint model to SMT-solvers, which accept nested expression trees as input.
For CP-solvers, however, further \emph{flattening} of the expression tree is required, as CP-solvers only support \emph{flat} constraints or their reification (\Cref{sec:flatten}).
That is, constraints in which each argument is a variable or a constant.
For ILP solvers, further linearization of the flattened constraints is required (\Cref{sec:linearize}).
Indeed, these solvers only support linear inequality constraints, so all logical constraints should be linearized as such.
Pseudo-Boolean (PB) solvers also support linear inequality constraints, but only over Boolean variables.
Hence, all integer variables must be encoded with an appropriate Boolean encoding.
\Cref{sec:int2bool} describes which encodings are currently implemented in our framework.
Finally, SAT-solvers only support \emph{clauses} of Boolean literals.
Hence, the set of constraints suited for PB-solvers must be processed further before it can be posted to a SAT-solver.
For this final step, our framework largely relies on external libraries such as \pypblib~\cite{pblib} and \pindakaas~\cite{pindakaas}.

Note that, given a target solver to transform the input model to, certain steps in the transformation may be skipped.
E.g., if a CP-solver supports the \talldiff constraint, then naturally it should not be decomposed by the transformation stack.
Similarly, some ILP solvers support the use of an implication constraint in their API; this means the linearization of implications is skipped in our transformation stack.
In practice, we implement this using an ``exclude'' set, which lists the operators that are left unchanged during a particular transformation.

Two key challenges for transforming a high-level constraint model will be to 1) correctly handle negation of arbitrary Boolean expressions and 2) avoid introducing superfluous auxiliary variables.
The first challenge will largely be handled during safening of partial functions, and subsequent elimination of negation constraints.
To avoid introducing unnecessary auxiliary variables, we apply Common Subexpression Elimination (CSE)~\cite{rendl2008eliminating} during flattening and subsequent transformations.
This ensures all semantically equivalent subexpressions are channeled to the same auxiliary variable.

For constraint optimization problems, the objective function may also need to be reformulated given a particular solver.
For example, several CP-solvers only support minimizing the value of a single variable, while ILP-solvers only support weighted linear objective functions.
Therefore, we apply the same reformulations to the numerical objective function as we do for the constraints, before posting the objective function to a solver.
The architecture of every transformation is designed such that it can take both Boolean and numerical expressions as input.

\subsection{Eliminating Partial Functions}
\label{sec:safening}

Our constraint modeling language allows the use of global functions.
These functions map a set of ``input'' variables to a single output variable.
However, some of the global functions implemented in our system are ``partial'' functions, for which the output can be \emph{undefined} (see \Cref{def:partial}).
Common examples of partial function constraints in constraint programming models are the \emph{division} constraint ($\consdiv{x}{y} = z$) and the \telement constraint (written as $\element{\xvar}{i} = v$).
Indeed, division is undefined when $(y = 0)$ and an array lookup is undefined when the index is out-of-bounds (i.e., $i < 1$ or $i > |\xvar|$).
When the domains of the variables in a partial function allow it to be undefined, we say the function is \emph{unsafe}, and it should be \emph{safened}.
We handle undefinedness using the \emph{relational semantics} for constraint modeling languages~\cite{frisch2009proper}.
That is, when the result of a numerical function is \emph{undefined}, it asserts its \emph{nearest Boolean context} to \tfalse.
In the context of our modeling language, this means the nearest Boolean parent expression is asserted to be \tfalse.
Both MiniZinc and Savile Row also implement the relational semantics for handling partial functions~\cite{stuckey2013minizincwithfunctions,nightingale2022savilerow}.

In \Cref{fig:partials} we show different expression trees that contain a partial function, in this case \tdiv.
We focus on partial functions for which the nearest Boolean context is \emph{not} top-level (e.g., \Cref{fig:nesteddiv,fig:negateddiv}).
Indeed, solvers supporting top-level partial function constraints will exclude any \emph{unsafe} value from the domains themselves, e.g., through root-level propagation for CP-solvers.

\input{images/partials}
\begin{figure}[t]
\centering
\begin{subfigure}[t]{0.25\textwidth}
\vspace{0pt}
\centering
\topleveldiv
\caption{Top-level}
\label{fig:top-leveldiv}
\end{subfigure}%
\begin{subfigure}[t]{0.25\textwidth}
\vspace{0pt}
\centering
\nestedsumdiv
\caption{Nested in $\textgrammar{IntExpr}$}
\label{fig:nestedsumdiv}
\end{subfigure}%
\begin{subfigure}[t]{0.25\textwidth}
\vspace{0pt}
\centering
\nesteddiv
\caption{Nested in a disjunction}
\label{fig:nesteddiv}
\end{subfigure}%
\begin{subfigure}[t]{0.25\textwidth}
\vspace{0pt}
\centering
\negateddiv
\caption{Nested in a negation}
\label{fig:negateddiv}
\end{subfigure}
\caption{
In \Cref{fig:top-leveldiv}, the nearest Boolean context of the division is the top-level \tequals constraint, so the model forbids $y \mapsto 0$. The same holds for \Cref{fig:nestedsumdiv}, where the only intermediate parent is \tsum (numerical). In \Cref{fig:nesteddiv}, the nearest Boolean context is the comparison within the top-level disjunction, so $y \mapsto 0$ causes the left-hand side to evaluate to \tfalse, requiring $p$ to be true. In \Cref{fig:negateddiv}, the nearest Boolean context is the negation, allowing solutions where the division is false or where $y \mapsto 0$.
}
\label{fig:partials}
\end{figure}

\subsubsection{Safening partial functions using guards} \label{sec:safening-how}

The general idea of safening a nested partial function is to introduce a \emph{guard}, which determines whether the partial function is defined or not, and to replace the argument that can cause the function to be undefined with a \emph{safe} alternative.
This guard is placed in conjunction with the nearest Boolean context~\cite{frisch2009proper}.
We observe that for our modeling framework, two kinds of guards can be needed, depending on the domain for which the partial function is defined.
Both cases are discussed below.

\paragraph{Case 1: Safening to a range of values}
The first type of partial functions is only defined for a contiguous range of values.
We will use the array lookup function $\element{X}{y}$ as an example, which is only defined when the index $y \in \range{1}{|X|}$. 
Hence, a natural guard is $(1 \leq y \leq |X|)$.
To safen the constraint, we introduce a new integer variable $y_\mathit{safe} \in \range{1}{|X|}$, and we ensure the original decision variable $y$ is equal to $y_\mathit{safe}$ whenever the function is defined.

\begin{example}[Safening of an \telement constraint]
Consider the constraint $(\element{X}{i} = v) \vee p$.
After safening, this constraint is reformulated as a conjunction of the constraints:
\begin{align}
    \big((\element{X}{y_\mathit{safe}} = v) \wedge (1 \leq y \leq |X|)\big) \vee p \quad& \wedge \label{example:cons:orig2}\\
    (1 \leq y \leq |X|) \implies (y = y_\mathit{safe}) \quad& \wedge \label{example:cons:guard} \\
    y_\mathit{safe} \in \range{1}{|X|} \quad&
\end{align}
where constraint \ref{example:cons:orig2} is the original, safe constraint and \ref{example:cons:guard} ensures the new index is equal to the original index variable whenever it is in the bounds of the array.
\end{example}

\paragraph{Case 2: Excluding a single value from the domain}
The second type of partial functions are undefined for a single value in the domain of one of their arguments.
Common examples are the $\consdiv{x}{y}$ and $\consmod{x}{y}$ functions, which are both undefined when $y = 0$. 
A natural guard is $y \neq 0$.
As our modeling language only allows integer variables with a contiguous domain (see \Cref{gra:cpmpy}), we need to model both sides of the domain around the excluded value separately.
More specifically, we introduce two auxiliary variables $y_\mathit{neg}$ and $y_\mathit{pos}$ to represent the safe negative and positive parts of the domain of $y$, respectively.

\begin{example}[Safening a nested division constraint]
Consider the unsafe constraint $(\consdiv{x}{y} = v) \vee p$ as introduced previously.
After safening, this expression becomes a conjunction:
\begin{align}
    \big((z = v) \wedge (y \neq 0)\big) \vee p \quad& \wedge \label{example:cons:orig}\\
    (y < 0) \implies (\consdiv{x}{y_\mathit{neg}} = z \wedge y = y_\mathit{neg}) \quad& \wedge \label{example:cons:low}\\
    (y > 0) \implies (\consdiv{x}{y_\mathit{pos}} = z \wedge y = y_\mathit{pos}) \quad& \wedge \label{example:cons:high} \\
    y_\mathit{neg} \in \range{\lb{y}}{{-1}} \wedge y_\mathit{pos} \in \range{1}{\ub{y}} \quad&
\end{align}
Where constraint \ref{example:cons:orig} is the original, but now \emph{safe} constraint; and constraints \ref{example:cons:low} and \ref{example:cons:high} ensure the auxiliary variables are asserted to the original $y$ when $y$ takes a value that is different from 0; and that the division constraint holds.
\end{example}
    
\subsection{Eliminating negation}\label{sec:negation}

After eliminating partial functions, we focus on eliminating \emph{negation} from the expression tree.
More specifically, we traverse the expression tree and ``push down'' any $\neg$ operator into its arguments.
For this, we follow a set of standard simplification rules.
To rewrite the negation of a comparison, we can simply replace the comparison with its negative counterpart.
That is, $=$ is mapped to $\neq$, $\leq$ to $>$ and $<$ to $\geq$.
For logical operators, we follow De Morgan's laws~\cite{demorgan1864}: $\neg(a \wedge b)$ is simplified to $(\neg a) \vee (\neg b)$, $\neg(a \vee b)$ to $(\neg a) \wedge (\neg b)$, and $\neg(a \implies b)$ to $a \wedge (\neg b)$.
When $a$ or $b$ is an expression, negation is further eliminated recursively.

Finally, negating global constraints is non-trivial, and no generic system exists for (efficiently) negating them.
However, for some global constraints, it is possible to define their negation in terms of other global constraints, as observed by \citet{fages2012reifying}.
Below is a non-exhaustive list of constraint reformulations for negated global constraints as implemented in our system:
\begin{itemize}
    \item $\neg\constable{X,T}$ is mapped to $\negtable{X,T}$
    \item $\neg\alldiff{X}$ is mapped to $\nvalue{X} < |X|$
    \item $\neg\regular{X,N,T,s,A}$ is mapped to $\regular{X,N,T,s,N \setminus A}$ for complete automata
    \item $\neg\cons{Xor}{x_1,x_2,\dots,x_n}$ is mapped to $\cons{Xor}{\neg x_1, x_2, \dots, x_n}$
\end{itemize}

For global constraints where no specialized negation is implemented, we keep the negated global constraint in the expression tree.
This is shown in \Cref{gram:negation}, where all negation operators are eliminated, apart from negated Boolean variables and negated global constraints.

Note that no transformations further down the transformation stack are allowed to explicitly introduce the negation of an expression.
E.g., when the decomposition of a global constraint requires negating one of its arguments, that negation operator must immediately be eliminated using the transformation described in this section.
For example, the decomposition of $\cons{IfThen}{x \leq y, z = 2, z = 3}$ is $\set{(x \leq y) \implies (z = 2), \neg(x \leq y) \implies (z = 3))}$. 
The latter of these constraints is then further rewritten as $(x > y) \implies (z = 3)$ to eliminate the negation operator from the new expression.

\input{grammars/negation.tex}

\subsection{Decomposing unsupported global constraints and functions}
\label{sec:decompose}

Our constraint modeling language supports the use of \emph{global constraints}~\cite{hoeve2006global,globalconstraintcatalog} to model constraint satisfaction or optimization problems.
However, not all solvers support (all) global constraints.
Indeed, while most CP-solvers support a variety of global constraints and functions, SMT or ILP solvers rarely support them (apart from some common functions such as \tmin or \tabs).
Hence, for all solvers, the unsupported global constraints and functions must be decomposed before posting the model to the solver.

Note that in our language, global constraints can be used as any other Boolean expression, and thus they can occur as a nested expression in the model.
However, CP-solvers typically do not support such arbitrary nesting of global constraints.
Instead, they only support using the global constraints at the ``top-level'' of the constraint model, where the solver will maintain its satisfaction during search.
Hence, we define two general types of ``supportedness'' of global constraints: they are either supported at the top-level, or supported as a nested expression.
Each solver interface is configured using two sets, \supported and \supportednested, that list the set of global constraints supported at the top-level and in a nested context, respectively.

In this transformation, we avoid decomposing the same global constraint or global function multiple times by keeping a cache of decomposed expressions.
This generally reduces the number of expressions and auxiliary variables in the decomposed constraint model.
While it may be uncommon for a user to write the exact same global constraint twice, decompositions or other reformulations may introduce the same expression in various places.

\begin{subsubsection}{Decomposing top-level global constraints\label{sec:decomp:global:tl}}

Each solver will support a subset of global constraints that are supported at the ``top-level'' of the constraint model.
This is the most common case where global constraints occur: e.g., \talldiff constraints in allocation problems or \tcumulative constraints in many scheduling problems.
Any top-level global constraint that is not listed in \supported, will need to be decomposed into a conjunction of expressions that is equivalent to the original global constraint.
For example, given a constraint solver that does not support the constraint $\alldiff{x,y,z}$, we can replace it with the conjunction $(x \neq y) \wedge (x \neq z) \wedge (y \neq z)$.
However, note that global constraints may be decomposed into other global constraints.
This means that we need to recursively check whether all unsupported global constraints and functions are eliminated.
An example is the decomposition of the \tgcc constraint, as shown in \Cref{fig:gcc-decomp}, which is decomposed using \tcount functions.
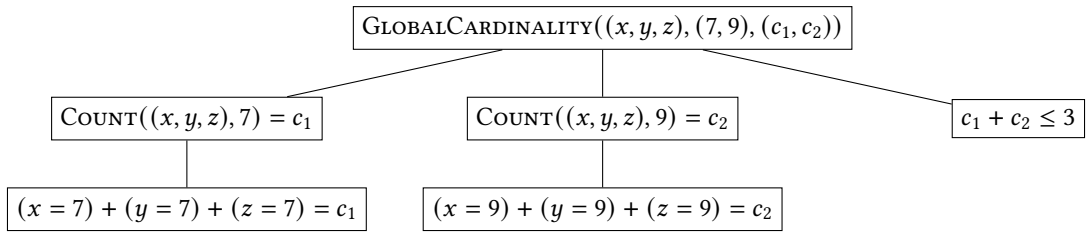
\begin{figure}[h]
    \centering
  \begin{tikzpicture}
  [level distance=12mm,level/.style={sibling distance=55mm/#1}]
  \node [draw,rectangle] {$\gcc{(x,y,z),(7,9),(c_1,c_2)}$}
    child {node[draw,rectangle] {$\cons{Count}{(x,y,z),7} = c_1$}
        child {node[draw,rectangle] {$(x = 7) + (y = 7) + (z = 7) = c_1$}}
    }
    child {node[draw,rectangle] {$\cons{Count}{(x,y,z),9} = c_2$}
        child {node[draw,rectangle] {$(x = 9) + (y = 9) + (z = 9) = c_2$}}
    }
    child {node[draw,rectangle] {$c_1 + c_2 \leq 3$}};
\end{tikzpicture}
\caption{Decomposing the \term{GlobalCardinalityCount} constraint into two global functions, which are in turn decomposed into nested arithmetic constraints.}
\label{fig:gcc-decomp}
\end{figure}

As we will discuss in this section, some global constraints require the introduction of \emph{auxiliary variables} to formulate the relation represented by the constraint.
Following \citet{beldiceanu2013reification}, our framework returns two sets of constraints as the decomposition of a global constraint $\globx$: a set of \emph{defining} constraints $\definingcons{\variables}{\auxvars}$ and a set of \emph{value} constraint(s) $\valuecons{\variables}{\auxvars}$.
The defining constraints determine the value of the auxiliary variables $\auxvars$, using the decision variables $\variables$.
The value constraints represent the truth value of the global constraint.
Note that this decomposition scheme is compatible with decomposing global functions as well. 
Indeed, while for global constraints, the value constraint is a set of constraints representing a single Boolean-valued conjunction, for global functions it is an integer-valued expression.

\begin{example}[Decomposition of \tmin using an auxiliary variable]
Consider the constraint $\consmin{x,y,z} \geq 3$.
If the solver does not support the global function \tmin, we decompose it using an auxiliary variable $m$.
More specifically, the defining constraints become 
$\definingcons{\set{x,y,z}}{\set{m}} = \big( m \leq x \big) \wedge \big( m \leq y \big) \wedge \big( m \leq z \big) \wedge \big( (m \geq x) \vee (m \geq y) \vee (m \geq z) \big)$
and the value expression is the auxiliary variable: $\valuecons{\set{x,y,z}}{\set{m}} = m$. As part of the decomposition procedure, the value expression $m$ is put in place of the original \tmin constraint: $m \geq 3$.
\end{example}

\begin{example}[Decomposition of \tmdd using auxiliary variables] \label{example:mdd}
Consider the following $\tmdd$ constraint: $\mddcons{(x,y),(S,1,A),(S,2,B),(A,1,F),(B,1,F),(B,2,F), F}$ with starting node $S$ and accepting state $F$.
This constraint is satisfied if the paths defined by the transitions of the MDD represent an accepting path through the MDD.
The decomposition uses a Boolean edge variable $\edgevarval{X}{Y}{i}$ for each edge $X \rightarrow Y$ with label $i$ in the MDD; and enforces a ``flow'' to the accepting state \cite{ahuja1993network}. Its defining and value constraints are:

\begin{minipage}{0.6\textwidth}
\begin{align*}
    E = &~\set{
    \edgevarval{S}{A}{1},
    \edgevarval{S}{B}{2},
    \edgevarval{A}{F}{1},
    \edgevarval{B}{F}{1}, 
    \edgevarval{B}{F}{2}} \\
    \definingcons{\set{x,y}}{E} =
    &~\big( (x = 1) \reifies \edgevarval{S}{A}{1} \big) \wedge
    \big( (x = 2) \reifies \edgevarval{S}{B}{2} \big) \wedge \\
    &~\big( (y = 1) \reifies \edgevarval{A}{F}{1} \vee \edgevarval{B}{F}{1} \big) \wedge
    \big( (y = 2) \reifies \edgevarval{B}{F}{2} \big) \wedge \\
    &~\big( \edgevarval{S}{A}{1} = \edgevarval{A}{F}{1} \big) \wedge \\
    &~\big( \edgevarval{S}{B}{2} = \edgevarval{B}{F}{1} + \edgevarval{B}{F}{2} \big)
\\
    \valuecons{\set{x,y}}{E} =
        &~\big( \edgevarval{S}{A}{1} + \edgevarval{S}{B}{2} = 1 \big) \wedge \\
        &~\big( \edgevarval{A}{F}{1} + \edgevarval{B}{F}{1} + \edgevarval{B}{F}{2} = 1 \big)
\end{align*}
\end{minipage}\vspace{5pt}%
\begin{minipage}{0.35\textwidth}
\hfill
\begin{tikzpicture}[
    >=Stealth,
    vertex/.style={draw,circle,minimum size=8mm,inner sep=0pt},
    label/.style={minimum size=8mm,inner sep=0pt}
]

\node[vertex] (S) at (0,1.5) {$S$};
\node[vertex] (A) at (-1,0) {$A$};
\node[vertex] (B) at (1,0) {$B$};
\node[vertex,double,double distance=1pt] (F) at (0,-1.5) {$F$};

\node[label] (X) at (2,0.80) {$x$};
\node[label] (Y) at (2,-0.80) {$y$};

\draw[->] (S) -- node[left] {1} (A);
\draw[->] (S) -- node[right] {2} (B);
\draw[->] (A) -- node[left] {1} (F);
\draw[->] (B) to[bend left=20] node[right] {1} (F);
\draw[->] (B) to[bend right=20] node[left] {2} (F);

\end{tikzpicture}
\end{minipage}

Here, the channeling constraints determine the value of the auxiliary variables and the consistency of flow are the defining constraints, and the constraints enforcing that there is a non-zero flow from the starting node to the accepting state are the value constraints.
\end{example}
\end{subsubsection}

While the auxiliary variables cannot be avoided in the previous two examples to reformulate the constraint, for many decompositions, we can circumvent explicitly introducing auxiliary variables in the decomposition.

\paragraph{Avoiding introduction of functionally defined auxiliary variables}
\label{sec:auxvars}

For many decompositions, the auxiliary variables used in the decomposition are functionally defined by the user-defined decision variables.
In general, we try to avoid explicitly introducing auxiliary variables that represent a subexpression used in the decomposition.
Instead, we use the subexpression itself as an argument to build the complex expression that represents the decomposition.
We will call this the \emph{functional form} of decomposing global constraints.
\begin{example}[Functional decomposition of \targmin]
\label{example:funcargmin}
Consider the following naive decomposition of the constraint $\argmin{X,i}$:
$(X[i] = w) \wedge (\consmin{X} = v) \wedge (w = v)$~\cite{beldiceanu2012reification}.
This decomposition introduces two auxiliary variables $w$ and $v$, to define the relation.
However, their value is functionally defined.
An equivalent reformulation which avoids the explicit use of auxiliary variables is $(X[i] = \consmin{X})$.
\end{example}

Solvers that support nested expressions will benefit from an explicitly nested structure. 
For other solvers, the flattening will introduce the auxiliary variables automatically, but it will use Common Subexpression Elimination (CSE) to avoid creating duplicates and will compute tight bounds on the variables automatically.
This scheme is also compatible with the decomposition of global functions, as the following example shows.

\begin{example}[Decomposing \tnvalue]
\label{example:funcnvalue}
The standard decomposition for the constraint $\nvalue{x,y,z} = n$ introduces a Boolean variable $\directencB{v}{i}$ for each of the domain values $x,y$ and $z$~\cite{bessiere2010nvalue}.
The decomposition then enforces the constraints: 
$\forall v \in \set{x,y,z} \forall i \in \domainof{v} : \directencB{v}{i} \reifies (v = i)$ and
$n = \sum_{i \in \domainof{x} \cup \domainof{y} \cup \domainof{z}} 
(\directencB{x}{i} \vee \directencB{y}{i} \vee \directencB{z}{i})$.
However, it is easier to formulate the decomposition over the subexpressions directly, which avoids explicitly introducing any auxiliary variables at this stage:
$n = \sum_{i \in \domainof{x} \cup \domainof{y} \cup \domainof{z}} ((x = i) \vee (y = i) \vee (z = i))$.
\end{example}

Avoiding the introduction of auxiliary variables at decomposition time has several advantages.
First, it reduces the overall number of variables for solvers that support nested expressions (see \Cref{sec:flatten}).
Indeed, some solvers (e.g., most SMT-solvers) directly support the nested constraints as shown in the example.

Secondly, given the \targmin constraint from \Cref{example:funcargmin}, it is not unthinkable that the user will use the expression $X[i]$ elsewhere in the constraint model -- say in constraint $C_2$.
If at some point in the transformation stack, an auxiliary variable is made for this other occurrence in the model (e.g., during flattening of $C_2$), then it can be used to replace the subexpression in the decomposition as well.
Hence, by postponing the introduction of auxiliary variables as long as possible in the transformation stack, we can localize all CSE-logic into the flattening transformation and potentially enable more opportunities to do CSE~\cite{rendl2008eliminating}.

\paragraph{ILP-friendly decompositions}
When reformulating a CSP for an Integer Linear Programming (ILP) solver, virtually all global constraints in the model must be decomposed.
Indeed, apart from occasional support for numerical global functions such as \tmin or \tabs constraints, ILP solvers typically only support linear inequalities as input language.
Standard ``CP-style'' decompositions of global constraints are often unsuited for ILP solvers.
A prime example is the \talldiff constraint, whose standard decomposition is a set of $n(n-1)$ binary disequality constraints~\cite{hoeve2001alldifferent}.
However, to linearize disequality constraints, we need to use big-M style constraints (see \Cref{sec:linearize}), which are generally unwanted for performant ILP models.
\citet{belov2016improved} proposes that many global constraints can instead be decomposed in an ``ILP-friendly'' way.
These specialized decompositions (implicitly) use a Boolean encoding variable $\directencB{x}{v}$ to represent that a categorical variable $x$ is assigned to value $v$.
Below, we show several examples of such decompositions.

\begin{example}[ILP-friendly flow decomposition of \talldiff]
Given the constraint $\alldiff{X}$ with a set of integer variables $X$, the flow-based decomposition is $\bigwedge_{v\in {\mathit{lb}..\mathit{ub}}} \sum_{x_i \in X}(x_i = v) \leq 1$
\end{example}

\begin{example}[ILP-friendly decomposition of \telement]
Given an array $A$ of integer \emph{constants}, and an index variable $i$ the constraint $A[i] = v$ can be decomposed as the linear constraint
$\sum_{j=1..|A|} a_j \times (i = j) = v$. 
The linear relaxation of this constraint ensures the bounds of variable $v$ are $[\mathit{min}(A), \mathit{max}(A)]$.
\end{example}

Our framework currently implements specialized linear decompositions for \talldiff, \telement, \ttable, \tregular and \tcircuit constraints.
Interestingly, by avoiding explicitly creating auxiliary variables when decomposing global constraints, several CP-style decompositions already work well for ILP solvers, as they are further processed by the \emph{linearize} transformation down the transformation stack.
For example, we decompose the function $\conscount{X,v}$ as $\sum_{x \in X} (x = v)$ for all solvers.
When targeting a CP-solver, our transformation pipeline will introduce Boolean variables $B_{x =v}$ used in $\sum_{x \in X} B_{x = v}$ and reification constraints $B_{x =v} \reifies (x = v)$.
For ILP solvers, the decomposition will be the same $\sum_{x \in X} \directencB{x}{v}$ but the direct encodings $\directencB{x}{v}$ can be done in a more ILP-friendly way.
\Cref{sec:linearize} further elaborates on the implementation of this feature.

\subsubsection{Decomposing nested global constraints}
We now focus on the case of fully reified global constraints, but the techniques are valid for any nesting of global constraints in \Cref{gra:cpmpy}.
When introducing auxiliary variables to decompose a \textit{nested} global constraint, we need to ensure their value is at all times functionally defined by the decision variables~\cite{beldiceanu2013reification}.
That is, given \emph{any} value to the decision variables, there is \emph{exactly one} satisfying assignment to the auxiliary variables.
We say the auxiliary variables are totally defined by the decision variables.
In practice, this means the defining constraints $\definingcons{\variables}{\auxvars}$ must represent a total function, written as $\definingtotal{\variables}{\auxvars}$, which can and should be enforced at the top-level of the constraint model:
$$
    \bv \reifies \glob{\variables} \quad \equiv \quad \definingtotal{\variables}{\auxvars} \wedge \bv \reifies \valuecons{\variables}{\auxvars}
$$

Indeed, enforcing the definition of auxiliary variables as part of the nested expression is not valid, as shown in the following example.

\begin{example}[Invalid decomposition of a reified constraint using auxiliary variables]
Consider the constraint ${\bv \reifies \conscount{(x,y,z), 1} \leq 2}$, which enforces that the number of occurrences of the value $1$ is at most two, if and only if $\bv$ is true.
An \emph{invalid} decomposition would be $\bv \reifies ((x = 1) + (y = 1) + (z = 1) = c \wedge c \leq 2)$.
Indeed, the assignment $\set{\bv \mapsto \tfalse, x \mapsto 1, y \mapsto 2, z \mapsto 3, c \mapsto 3}$ satisfies the reformulation, but clearly does not satisfy the original reified constraint:
the Boolean reification variable is \tfalse, but the original comparison with the \tcount constraint on the right-hand side is satisfied by the assignment.
\end{example}

For this example of the \tcount constraint, it is easy to see that the auxiliary variable $c$ is functionally defined by the variables $x,y$, and $z$.
However, for other decompositions, this is less trivial, as the value of the auxiliary variables is defined through several constraint, as shown in the following example of a reified \tmdd constraint.

\newcommand{\extracol}{blue}
\newcommand{\extra}[1]{\textcolor{\extracol}{#1}}
\begin{example}[Decomposition of a reified \tmdd constraint] \label{example:mdd-reif}

Consider again the constraint from \Cref{example:mdd}, but now as a reification:
$\bv \reifies \mddcons{(x,y),(S,1,A),(S,2,B),(A,1,F),(B,1,F),(B,2,F), F}$.
Our previous decomposition is not suited to decompose this reified \tmdd constraint.
Indeed, the assignment
$\set{
\bv \mapsto \tfalse,
x \mapsto 3,
y \mapsto 2
}$
is not allowed by the decomposition introduced in the previous example,  
as it enforces variables $x$ and $y$ to take either value 1 or 2.
To circumvent this issue, the MDD must be expanded with a ``garbage state'' G for each layer in the MDD.
This allows to include all remaining domain values from each of the nodes.
For $\domainof{x} = \domainof{y} = \range{1}{3}$ we get:
\begin{minipage}{0.69\textwidth}
\begin{align*}
    E = \set{&\edgevarval{S}{A}{1},
             \edgevarval{S}{B}{2},
             \edgevarval{A}{F}{1},
             \edgevarval{B}{F}{1},
             \edgevarval{B}{F}{2},\\
             &\extra{\edgevarval{S}{G_x}{3}},
             \extra{\edgevarval{A}{G_y}{2},\edgevarval{A}{G_y}{3}},
             \extra{\edgevarval{B}{G_y}{3}}, \\
             &\extra{\edgevarval{G_x}{G_y}{1},\edgevarval{G_x}{G_y}{2},\edgevarval{G_x}{G_y}{3}}
             }\\
    \definingtotal{\set{x,y}}{E} =
    &\big( (x = 1) \reifies \edgevarval{S}{A}{1} \big) \wedge \\
    &\big( (x = 2) \reifies \edgevarval{S}{B}{2} \big) \wedge \\
    &\extra{
    \big( (x = 3) \reifies \edgevarval{S}{G_x}{3} \big)
    } \wedge
    \\
    &\big( (y = 1) \reifies \edgevarval{A}{F}{1} \vee \edgevarval{B}{F}{1} \vee \extra{\edgevarval{G_x}{G_y}{1}} \big) \wedge \\
    &\big( (y = 2) \reifies \edgevarval{B}{F}{2}  \vee \extra{\edgevarval{A}{G_y}{2} \vee \edgevarval{G_x}{G_y}{2}} \big) \wedge \\
    &\extra{\big( (y = 3) \reifies \edgevarval{B}{G_y}{3} \vee \edgevarval{A}{G_y}{3} \vee \edgevarval{G_x}{G_y}{3} \big)} \wedge \\
    &\big( \edgevarval{S}{A}{1} = \edgevarval{A}{F}{1} + \extra{\edgevarval{A}{G_y}{2} + \edgevarval{A}{G_y}{3}} \big) \wedge \\
    &\big( \edgevarval{S}{B}{2} = \edgevarval{B}{F}{1} + \edgevarval{B}{F}{2} + \extra{\edgevarval{B}{G_y}{3}} \big) \wedge \\
    &\extra{\big( \edgevarval{S}{G_x}{3} = \edgevarval{G_x}{G_y}{1} +\edgevarval{G_x}{G_y}{2} + \edgevarval{G_x}{G_y}{3} \big)}
\end{align*}    
\end{minipage}\hfill%
\begin{minipage}{0.3\textwidth}
\begin{tikzpicture}[
    >=Stealth,
    vertex/.style={draw,circle,minimum size=8mm,inner sep=0pt}
]
\node[vertex] (S) at (0,2) {$S$};
\node[vertex] (A) at (-1,0) {$A$};
\node[vertex] (B) at (1,0) {$B$};
\node[vertex,\extracol] (Gx) at (-2,0) {$G_x$};
\node[vertex,double,double distance=1pt] (F) at (0,-2) {$F$};
\node[vertex,\extracol] (Gy) at (-2,-2) {$G_y$};

\node[label] (X) at (2,1) {$x$};
\node[label] (Y) at (2,-1) {$y$};

\draw[->] (S) -- node[left] {1} (A);
\draw[->] (S) -- node[right] {2} (B);
\draw[->] (A) -- node[right] {1} (F);
\draw[->] (B) -- node[right] {1,2} (F);

\draw[->,\extracol]
    (S) to[bend right] node[left] {3} (Gx);

\draw[->,\extracol]
    (Gx) to[bend right=20] node[right] {1,2,3} (Gy);

\draw[->,\extracol]
    (A) to[bend left=20] node[right] {2,3} (Gy);

\draw[->,\extracol]
    (B) to[bend left, looseness=1.4, in=110, out=75] node[right=2pt] {3} (Gy);

\end{tikzpicture}
\end{minipage}

Now, all edge variables are uniquely defined given any value to the decision variables $x$ and $y$. 
The value constraints can be kept as before, as they enforce a flow matching the MDD constraint. 
Indeed, if any blue edge is activated, the constraints below cannot be satisfied.
\begin{align*}
    \valuecons{\set{x,y}}{E} =
        &\big( \edgevarval{S}{A}{1} + \edgevarval{S}{B}{2} = 1 \big) \wedge \\
        &\big( \edgevarval{A}{F}{1} + \edgevarval{B}{F}{1} + \edgevarval{B}{F}{2} = 1 \big)
\end{align*}
\end{example}

Note that much of the complexity of decomposing reified or nested global constraints as described in related works~\cite{beldiceanu2012reification,beldiceanu2013reification,fages2012reifying} can be avoided by not introducing auxiliary variables at all.
Indeed, when the auxiliary variables can be defined directly using a simple expression, we adopt the functional form of writing the decomposition for a global constaint or function.
For solvers that do not accept nested expressions, the flattening transformation described in \Cref{sec:flatten} then handles all logic for introducing auxiliary variables and determining their value based on the decision variables.
The example below shows how such a functional decomposition can simplify nested decompositions.

\begin{example}[Decomposing reified \targmin constraint]
Consider the reified constraint $\bv \reifies (\argmin{X} = i)$.
Using the naive decomposition with two auxiliary variables, the decomposition of the reification becomes:
$(X[i] = v) \wedge (\consmin{X} = w) \wedge \big(\bv \reifies (v = w)\big)$ where the \emph{definition} of $v$ and $w$ is enforced at top-level.
However, by writing the decomposition functionally, we obtain $\bv \reifies (\consmin{X} = X[i])$, which makes the defining expression empty, and we can simply replace the nested global constraint with its (nested) value expression.
\end{example}

As described previously, we can use specialized decompositions for some global constraints when targeting an ILP solver.
More generally, we allow a solver to specify a set of ``custom'' decompositions, allowing solvers to deviate from the default decomposition for specific global constraints or functions.
However, some of these decompositions, e.g., the decomposition of \tcircuit by \citet{miller1960integer}, are only valid when the constraint is at the top-level of the constraint model, not when nested in another expression.
Hence, when a \tcircuit constraint occurs in a nested context, we always decompose it using its ``traditional CP'' decomposition using order variables instead.
We next describe how to detect and use the more efficient decomposition when possible.

\subsubsection{Decomposing global constraints in positive context}

As discussed previously, our system supports arbitrary nesting of global constraints.
We observe that for some nestings, the global constraint occurs in \emph{positive context} in the expression tree.
Common examples of such nesting are half-reified constraints~\cite{feydy2011half} or constraints occurring in a disjunction~\cite{jefferson2010implementing}.
In such cases, the global constraint is never \emph{enforced} to be false.
Hence, when decomposing a global constraint in a nested positive context, we can simply re-use the decomposition as if the constraint were posted top-level (\Cref{sec:decomp:global:tl}).
Clearly, this is advantageous for many global constraints, as it avoids the requirement that auxiliary variables are functionally defined, and this can avoid extra variables and/or constraints, as \Cref{example:mdd,example:mdd-reif} have shown.

\paragraph{Reformulation of half-reified global constraints}

As a final optimization to the decomposition transformation, we consider the case where a CP solver supports a global constraint at the top-level of the model, but not in a nested context.
When the global constraint occurs in a positive context, \citet{bleukx2026halfreified} observed that one can avoid decomposing the global constraint, and instead introduce a set of auxiliary variables and a channeling constraint that uses the original (efficient) global constraint in a non-nested way.
To illustrate this, consider the top-level constraint $\bv \implies \alldiff{x,y,z}$.
If the solver does not support the \talldiff constraint in a nested expression, but does support  the top-level \talldiff constraint (which is the case for most CP solvers), then we can reformulate the constraint as:
$\alldiff{x',y',z'} \wedge (\bv \implies (x = x' \wedge y = y' \wedge z = z'))$ where $x',y'$ and $z'$ are auxiliary variables.
When the Boolean variable $\bv$ is set to \ttrue, the reified channeling constraint ensures the auxiliary variables are equal to the original decision variables, and the global constraint is asserted to be \ttrue.
When the indicator variable is set to \tfalse, the auxiliary variables are \emph{free}, and the reformulation is valid as long as $\alldiff{x',y',z'}$ is satisfiable.
This approach can be implemented generically for any global constraint, with several optimizations possible for functional or total relation constraints~\cite{bleukx2026halfreified}.
Using the \supported and \supportednested configured for each solver, we can easily determine which global constraints to reformulate using this method.

\subsection{Flattening by substituting expressions with auxiliary variables}
\label{sec:flatten}

Many (CP) solvers do not accept nested expressions, but rather accept only \emph{flat} constraints.
That is, constraints where all arguments are either constants or variables (= leaf nodes in the expression tree).
This step of the transformation pipeline is responsible for eliminating any subexpressions from the arguments of primitive operators and global constraints.
This algorithm traverses the entire expression tree of the constraint model and substitutes any nested expression with a freshly introduced auxiliary variable.
Then, the auxiliary variable is asserted to be equal to the subexpression it replaces, which is implemented in the helper function $\call{GetOrMakeVar}$ shown in \Cref{algo:getormakevar}.
For Boolean subexpressions, this results in a new reification constraint; for integer subexpressions, an equality constraint links the auxiliary variable with the subexpression it replaces.
To avoid introducing unnecessary auxiliary variables, we implement two well-known optimizations: in-place simplification of operators and Common Subexpression Elimination (CSE) over normalized expressions.

An implementation with a CSE-map and simplify procedure is shown in \Cref{algo:flatten}, where $F$ collects the flat constraints and $C'$ those that have to be recursively flattened. 
The algorithm first tries to simplify the expression by checking its immediate subexpressions and merging/rewriting associative operators.
It then checks if it is already in flat normal form; it then normalizes reifications, comparisons, negated global constraints, and other Boolean expressions to the flat normal form described in \Cref{grammar:flat2}.
This normalization iterates over the arguments of a subexpression and replaces any subexpression with a fresh auxiliary variable.
The pseudocode for \call{NormalizeBoolExpr} and \call{NormalizeNumExpr} is shown in Appendix \ref{appendix:normalize}.

\newcommand{\getormakevar}[2]{\call{GetOrMakeVar}(#1,#2)}
\newcommand{\normalizeargs}[1]{\call{NormalizeArgs}(#1)}
\newcommand{\normalizedboolexpr}[2]{\call{NormalizeBoolExpr}(#1, #2)}
\newcommand{\normalizednumexpr}[2]{\call{NormalizeNumExpr}(#1, #2)}

\newcommand{\todolist}{C'}
\renewcommand{\expr}{\mm{\mathit{expr}}}
\input{algorithms/flatten}

\paragraph{In-place simplifications operators}

A first optimization is to detect associative n-ary operators ($+, -,\vee, \wedge$) being used as a subexpression to the same operator.
For example, when a \textsc{Sum} constraint uses another \textsc{Sum} as a subexpression, they can clearly be merged, without introducing an auxiliary variable.
While it is uncommon for users to explicitly write such nested sum constraints, they are often the result of previous transformations. Additional care is taken to turn a sum where at least one of the terms is a multiplication of a constant and a variable, into a weighted sum, as well as to merge nestings of weighted and unweighted sums into a single weighted sum.

\begin{example}[Nested \textsc{Sum} after decomposing \tcount]

Consider the user-constraint ${2a + \conscount{\set{x,y,z},1} \geq 3}$.
If the solver does not support the \tcount constraint, it will be decomposed in-place by the previous transformation.
This results in the expression $2a + ((x = 1) + (y = 1) + (z = 1)) \geq 3$.
The simplification method at line 4 in \Cref{algo:flatten} will then merge the left-hand side of the sum constraint to a single weighted sum, instead of introducing an auxiliary variable for the decomposition of \tcount.
\end{example}

Secondly, for Boolean operators $\implies$ and $\vee$ and $\wedge$, we also apply distributivity and rewriting rules when they occur together.
\Cref{tab:reformulation-bool} lists all such simplification rules implemented in our flattening transformation.

\begin{table}[h]
    \centering
    \caption{Reformulation rules for nested $\implies, \vee$ and $\wedge$ operators to avoid unnecessary auxiliary variables}
    \begin{tabular}{c|c}
        Input & Output \\ \hline
        $a \vee (b \implies c)$ & $a \vee \neg b \vee c$ \\
        $(a \vee b) \implies c$ & $(\neg c \implies \neg a) \wedge (\neg c \implies \neg b)$
    \end{tabular} \hspace{20pt}
    \begin{tabular}{c|c}
        Input & Output \\ \hline
        $(a \implies b) \implies c$ & $(\neg c \implies a) \wedge (\neg c \implies \neg b)$ \\
        $(a \wedge b) \implies c$ & $\neg a \vee \neg b \vee c$
    \end{tabular}
   
    \label{tab:reformulation-bool}
\end{table}

\paragraph{Common Subexpression Elimination}

In constraint programming models, it is common to use a particular subexpression in multiple places in the model.
During the flattening process, we generate new auxiliary variables for each subexpression we encounter.
To avoid making duplicate variables, we keep a cache of already introduced auxiliary variables in a ``Common Subexpression Elimination Map'', which we refer to as $\csemap$.
This map keeps track of which (sub)expression is related to which auxiliary variable.
Hence, whenever a new variable needs to be introduced, we first check whether a variable for the expression already exists in the map.
To avoid creating separate auxiliary variables for two expressions that are syntactically different but semantically equivalent, we \textit{normalize} expressions into a normal form. 
The two simplest normalizations are that a comparison between a constant and an expression always has the constant on the right-hand side; and a multiplication between a constant and an expression has the constant as first argument. 
The in-place reformulation for associative operators is also performed as part of the normalization. 
We also implement normalization of expressions of the form $\textgrammar{Intvar} \textgrammar{Comparison} \textgrammar{Constant}$.
That is, all comparisons of this form are converted to either $\textgrammar{Intvar} = \textgrammar{Constant}$ or $\textgrammar{Intvar} \geq \textgrammar{Constant}$.
For example, when flattening $(x \neq 3) + y \leq 5$, we introduce a new Boolean variable to represent $x = 3$, say $\directencB{x}{3}$, and reformulate the constraint to $\neg \directencB{x}{3} + y \leq 5$.
This approach also improves the linearization of categorical variables, as we will describe later in \Cref{sec:linearize}.

\input{algorithms/getormakevar}

$ $

\noindent
\Cref{algo:getormakevar} shows the pseudocode for our lightweight implementation of common subexpression elimination~\cite{rendl2008eliminating}.
The resulting output \ignore{of the flattening transformation} is described in \Cref{grammar:flat2}, and closely resembles the language most CP solvers accept in their API.
That is, we allow linear expressions, clauses, global constraints, and global functions in a comparison.
Note that each of these expressions can still occur in a reified context.

The following two sections describe two additional post-flattening transformations that solvers can choose to activate or not.

\input{grammars/flat-alt}

\subsubsection{Normalize global functions}
The flat normal form allows global functions to be used in arbitrary comparisons and in reified comparisons.
E.g., the constraint $\bv \reifies \max(x,y,z) \leq p$ is a valid flat constraint.
However, most CP-solvers don't directly support such constraints.
Indeed, reified global constraints are rarely supported, and global functions are typically not supported in arbitrary comparisons.
Instead, solvers implement a separate global constraint $\cons{MaxEqual}{(x,y,z), p}$, which forces $p$ to be the maximum value among variables $x,y,z$.
This transformation introduces a fresh auxiliary variable (again through \call{GetOrMakeVar}) for each global function that should be rewritten to a top-level equality constraint:
$$
    F(X) \gcomparison Y \quad\equiv\quad (F(X) = Y') \wedge (Y' \gcomparison Y) 
$$

and 
$$
    \bv \reifies F(X) \gcomparison Y \quad\equiv\quad (F(X) = Y') \wedge (\bv \reifies Y' \gcomparison Y) 
$$
This latter reformulation was first proposed by \citet{beldiceanu2013reification} as a reformulation for reified total function constraints.
Note that some non-equality comparisons are supported by CP-solvers for certain global functions.
For example, the constraint $\nvalue{X} \leq c$ maps to the global constraint $\atmostnvalue{X, c}$, which is supported by several CP-solvers.
Hence, the implementation for normalizing global functions is a separate modular transformation that accepts a list of global functions for which only equality is supported.
\Cref{grammar:equality} describes the resulting set of allowed constraints in case all global functions only support equality.

\input{grammars/only-numexpr-equality}

\subsubsection{Reification to two implications}
\label{sec:reification}

In this second post-flatten transformation, we transform all reification constraints into a conjunction of two half-reification constraints for solvers that only support the latter.
That is, for a reified constraint $\textgrammar{Lit} \reifies \textgrammar{PrimitiveExpr}$, we introduce two constraints $\textgrammar{Lit} \implies \textgrammar{PrimitiveExpr}$ and $\neg\textgrammar{Lit} \implies \neg\textgrammar{PrimitiveExpr}$.
Note that the negations in the second half-reification are immediately passed to the \emph{eliminate negation} transformation described in \Cref{sec:negation} to ensure everything remains in flat normal form.
Additionally, as unsupported nested global constraints have already been decomposed before, introducing the negation of a supported nested global constraint at this stage in the transformation stack does not cause any issues, as the downstream solver supports nested (and hence also negated) global constraints (e.g., \choco~\cite{prudhomme2016choco}).
\Cref{grammar:reification} describes the set of constraints allowed after this normalization step.
For ILP, PB, and SAT solvers, we pass a flag to this transformation that keeps fully reified comparisons between an integer variable and a constant.
More precisely, we allow $\textgrammar{Lit} \reifies \textgrammar{Intvar} \textgrammar{Comparison} \textgrammar{Int}$ constraints to remain.
This allows for detecting integer variable encodings, which will benefit the ``linearize'' transformation described in the next section.

\input{grammars/only-implies}

\subsection{Linearization}
\label{sec:linearize}
We now describe how to further transform a set of flat normal form expressions into a format suitable for ILP-solvers.
The output of this transformation phase will also serve as input for further processing required for PB and SAT solvers, which we describe in \Cref{sec:int2bool}

We will assume any global constraints not supported by the ILP solver at hand are already decomposed in previous stages, that all full-reification constraints (except for $\textgrammar{Var} \textgrammar{Comparison} \textgrammar{Constant}$) have been split up into half-reifications, and that supported global functions are only used in compatible comparisons.
E.g., ILP solvers like SCIP \cite{achterberg2008scip} and Gurobi \cite{url:gurobi} allow to post a $\tmax$-equality constraint and a limited set of other global operators and constraints.
Disjunctions are trivially linearized by transforming to \emph{greater than 1 comparison}, e.g., $x \vee y \vee z \equiv x + y +z \geq 1$.
Strictly less-than and greater-than comparisons are eliminated by offsetting the right-hand side of the integer comparison with +1 and -1, respectively.
This leaves us with disequality constraints between a linear expression and a constant, half-reified linear constraints, and fully-reified equals and not-equals comparisons between a variable and a constant.
In the following two subsections we discuss how to linearlize these remaining constraints.

\input{grammars/linear}

\subsubsection{Linearization of half-reified linear inequality constraints}
We first consider implication constraints of the form $\textgrammar{Lit} \implies \textgrammar{LinExpr} \geq \textgrammar{Int}$.
As a generalization for any linear expression, we will write such a constraint as $\bv \implies \sum_{i = 1..n} w_i x_i \geq v$ in this section.
Thus, we assume that the right-hand side of any half-reification in the output of the flat normal form in \Cref{grammar:flat2} is already linearized by rewriting unsupported comparisons or by rewriting disjunctions to a linear constraint, and we just focus on linearizing the half-reification itself.
To linearize this constraint, we rely on a big-M reformulation~\cite{dantzig1963linear}.
The main idea is to convert the expression above to 
$M\cdot \neg \bv + \sum_{i=1..n} w_i x_i \geq v$.
Here, $M$ is a large integer constant that ensures the inequality is satisfied when $b$ is \tfalse, regardless of the values assigned to decision variables $x_i$.
When $b$ is \ttrue, the first term of the sum on the left-hand side of the inequality is 0, and the right-hand side of the half-reification is asserted to hold.

\subsubsection{Linearization of disequality}

Constraints of the form $x \neq y$ are reformulated using two half-reification constraints.
That is, we linearize any disequality constraint by introducing an auxiliary Boolean variable $b$, and introduce the constraints $b \implies x - y \leq -1$ and $\neg b \implies x - y \geq 1$.
These constraints are then linearized using a big-M reformulation as described above.
This results in two big-M constraints, one for each side of the disequality.

\subsubsection{Encoding categorical integer variables} \label{sec:linearize:categorical}
During flattening, we introduce an auxiliary Boolean variable for any comparison that is used as a subexpression in a constraint.
E.g., after decomposing, $\conscount{(x,y,z),3} = 1$ to $(x = 3) + (y = 3) + (z = 3) = 1$, it is flattened to 
$\set{
a_1 + a_2 + a_3  = 1,
a_1 \reifies (x = 3),
a_2 \reifies (y = 3),
a_3 \reifies (z = 3)
}$.
Such structures are common when integer variables are used as categorical variables, and each integer value represents an item of an enumerated type~\cite{stuckey2022enumerated}.
Naively linearizing each of the reification constraints separately would introduce 4 big-M constraints for each reification.
Indeed, let's consider the first constraint $a_1 \reifies (x = 3)$.
This requires two big-M constraints to model the half-reifications $a_1 \implies x \leq 3$ and $a_1 \implies x \geq 3$, and another two big-M constraints, and an auxiliary variable to linearize $\neg a_1\implies x \neq 3$:
\begin{align*}
    a_1 \reifies (x = 3) \equiv \big(a_1 \implies (x \leq 3)\big) \wedge \big(a_1 \implies (x \geq 3)\big) \wedge \big(\neg{a_1} \implies (x \neq 3)\big)
\end{align*}
and
\begin{align*}
     a_1 \implies (x \leq 3) &\equiv M_1 \neg{a_1} + x \leq 3\\
     a_1 \implies (x \geq 3) &\equiv  M_2 \neg{a_1} + x \geq 3 \\
     \neg{a_1} \implies (x \neq 3) &\equiv \big(\neg{a_1} \implies (b \implies x \leq 2)\big) \wedge \big(\neg{a_1} \implies (\neg b \implies x \geq 4)\big) \\
                                   &\equiv \big(\neg{a_1} \implies (M_{3} \neg b + x \leq 2)\big) \wedge \big(\neg{a_1} \implies (M_{4} {b} + x \geq 4)\big) \\
                                   &\equiv \big(M_5 a_1 + M_{3} \neg b + x \leq 2\big) \wedge \big(M_6 a_1 + M_{4} {b} + x \geq 4\big) \\
\end{align*}

As these Big-M constraints can hurt performance in ILP solvers due to their weak linear relaxation, this is to be avoided.
Instead, we wish to detect and use the direct encoding for integer variables whenever a categorical integer variable is used.
That is, instead of linearizing the expression $a_1 \reifies (x = 3)$, we encode the variable $x$ as 
$1\cdot\directenc{x}{1} + 2\cdot\directenc{x}{2} + 3\cdot a_1 + 4\cdot\directenc{x}{4} + 5\cdot\directenc{x}{5} = x$.
Notice that we do not introduce another variable $\directenc{x}{3}$, but simply reuse the already introduced variable $a_1$ to represent that $x$ is equal to 3.
In practice, this involves iterating over the $\csemap$ and extracting all auxiliary Boolean variables that map to an expression of the form $\textgrammar{IntVar} = \textgrammar{Int}$. 
We allow the user to set the minimum number of values before the direct encoding is created; by default, this is set to 2.

We implement a similar optimization for integer variables that occur frequently in a $\leq$ comparison to a constant.
That is, given enough $\textgrammar{IntVar} \geq \textgrammar{Int}$ expressions for the same integer variable in the $\csemap$, we encode the integer variable using its order encoding (see \Cref{sec:int2bool}).
This again avoids linearizing reified inequality constraints with Big-M constraints when such inequalities are used in logical constraints.
An example of such a case is when using the time-decomposition of the \tcumulative constraint~\cite{feydy2009why}.

Currently, we eagerly select the direct encoding when for a single integer variable both $\textgrammar{IntVar} = \textgrammar{Int}$ and $\textgrammar{IntVar} \geq \textgrammar{Int}$ expressions are present.
Future optimization could include encoding the integer variable twice and channeling the encodings to achieve better solving times~\cite{bierlee2022coupling}.

$ $

\noindent Note that most ILP solvers do not support negated Boolean variables (i.e., negative literals).
Hence, as an additional transformation, we replace any literals $\neg b$ with $(1 - b)$.
This is trivial to do for linear constraints, as weighted terms of a sum can simply be multiplied by $-1$, and constants can be subtracted from the right-hand side of the comparisons.
E.g., $3 (\neg x) + 4y \geq 2$ can be rewritten to $3(1-x) + 4y \geq 2$ and normalized to $-3x + 4y \geq -1$.

\subsection{Encoding All Integer Variables as Boolean Literals}
\label{sec:int2bool}

To transform constraints in our language for (Pseudo-)Boolean solvers, all integer variables must be encoded into Boolean literals.
For this, we reuse most of the modular transformation stack as done for ILP solvers, without the final conversion of negative to positive literals.
We currently support the most common encodings: direct encoding~\cite{walsh2000sat}, order encoding~\cite{tamura2009compiling}, and log-encoding (also known as binary encoding)~\cite{ernst1997automatic}.
These are all implemented through a common interface with three main functionalities:
\begin{enumerate}
    \item Encoding a term of the form $\textgrammar{Int} \times \textgrammar{Var}$, 
    \item encoding a comparison of the form $\textgrammar{Var} \textgrammar{Comparison} \textgrammar{Int}$ and
    \item formulating the consistency constraints to ensure the encoding is valid.
\end{enumerate}

\paragraph{Direct encoding}
The direct encoding of integer variables introduces a Boolean literal for each of the values in the integer variable domain.
To encode a term $k \times x$, we simply return the weighted sum $k \times \sum_{i \in \domainof{x}}{(i \times \directenc{x}{i})}$
Encoding a $=$ or $\neq$ constraint is as trivial as returning the matching Boolean literal, to encode a $\leq$ or $\geq$ constraint, we construct a conjunction of negative literals of the excluded part of the domain.
To ensure the encoding is consistent, we add the constraint $\sum_{i \in \domainof{x}} \directenc{x}{i} = 1$.
This results in a total of $|\domainof{x}|$ literals.

\paragraph{Order encoding}
The order (or thermometer) encoding of an integer variable uses a Boolean ``inequality'' variable for each value in the domain.
That is, we introduce a literal $\orderenc{x}{i}$ to represent that $x$ is greater than or equal to the value $i$.
Similar to the direct encoding, encoding a weighted term involves simply weighting each literal in the encoding: $k \times (\lb{x} + \sum_{i \in \range{\lb{x}+1}{\ub{x}}}{\orderenc{x}{i}})$.
The encodings for $\leq$ and $\geq$ constraints are now a single literal, while $=$ and $\neq$ constraints are encoded by first rewriting them to combinations of $\leq$ and $\geq$ constraints.
To ensure consistency of the encoding, we require $\orderenc{x}{i} \implies \orderenc{x}{i-1}$ for all values $i$ in $\domainof{x}$.
This results in a total of $|\domainof{x}|-1$ literals.

\paragraph{Log encoding}
The log encoding for integer variables is based on the binary representation of the value assigned to the integer.
That is, we use a list of $m$ Boolean literals, where $\logenc{x}{i}$ is set to \ttrue if the offset binary representation of the value of $x$ is set to \ttrue \cite{soh2015hybrid}.
To encode a term, we construct the sum $\sum_{0..m-1} (k \times 2^i \times \logenc{x}{i}) + \lb{x}$.
Comparisons are encoded as-is, by using the above sum as the left-hand side in place of the integer variable.
For consistency, we ensure the encoding is always smaller than or equal to the upper bound of the integer variable.
This results in $\lceil\log_2(|\domainof{x}|)\rceil$ literals to encode $x$.

$ $

\noindent
The output of the encoding step is a set of pseudo-Boolean constraints.
Our modular encoding module allows for future improvements to this encoding step, for example, to use multiple encodings of the same integer variable depending on which types of constraints it is used in \cite{bierlee2022coupling,soh2015hybrid}.

\subsection{To Conjunctive Normal Form (CNF)}
For SAT solvers, which require CNF input, we reuse the modular transformations as for Pseudo-Boolean solvers, except that in the linearisation transformation we indicate that half-reified constraints are supported and hence need not be encoded with Big-M constraints.
After the transformation to Pseudo-Boolean, what is left to do is to rewrite the implications as disjunctions and to formulate the weighted linear constraints into CNF.
This step has been well studied~\cite{marquessilva2007towards,prestwich2021cnfencodings,bierlee2025revisiting,aavani2011translating}, and in our framework, we use external libraries that implement this, such as \pypblib~\cite{pblib} and  \pindakaas~\cite{pindakaas}. 
The to-CNF transformation allows indicating that unweighted sums (cardinality constraints) are supported and hence need not be encoded for SAT solvers that support this natively, such as Minicard~\cite{liffiton2012minicard}.

\section{Experimental results}

We use our implementation of the above transformations to investigate the following experimental questions.

\newlist{EQ}{enumerate}{2} %
\setlist[EQ,1]{label=\bfseries EQ\arabic*., leftmargin=30pt}
\setlist[EQ,2]{label=\bfseries EQ\arabic{EQi}.\arabic*., leftmargin=30pt}

\begin{EQ}
    \item How does the model for a finite domain constraint model change throughout the transformation stack?
    \item What is the impact of the transformation optimizations on solving time for ILP, PB and SAT solvers?
    \begin{EQ}
        \item What is the impact of using ILP-friendly decompositions for global constraints?
        \item What is the effect of encoding categorical integer variables with a direct Boolean encoding?
        \item What is the impact of using positive-only decompositions for global constraints?
    \end{EQ}
\end{EQ}

We consider all 250 instances of the \texttt{COP} track of the 2024 edition of the XCSP3 competition.
These optimization models consider a wide variety of constraints, including global constraints and functions.
The instances are parsed into CPMpy v1.0.0 using the built-in datasets and IO tools~\cite{sergeys2026unified}.

To answer the second experimental question, we run these benchmark instances on three different solvers integrated in the framework: MIP solver Gurobi v.13.0.2~\cite{url:gurobi}, Pseudo-Boolean solver Exact v2.3.0~\cite{exact}, and MaxSAT solver RC2 via PySAT v1.9.dev5~\cite{ignatiev2018pysat}.
All experiments were run on a single thread of an Intel(R) Xeon(R) Silver 4514Y and with a memory limit of 8GB.

\subsection{EQ1: effect of transformation stack on constraint model}

\begin{figure}[t]
    \centering
    \begin{minipage}[c]{0.19\linewidth}
    \includegraphics[width=\linewidth]{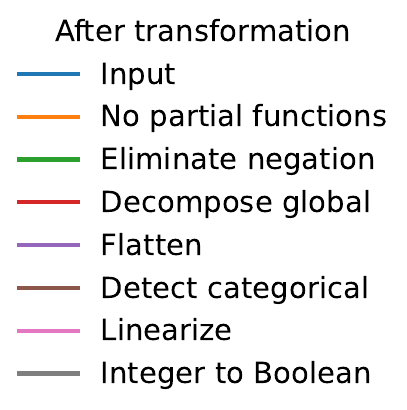}    
    \end{minipage}%
    \begin{minipage}[c]{0.81\linewidth}
    \includegraphics[width=\linewidth]{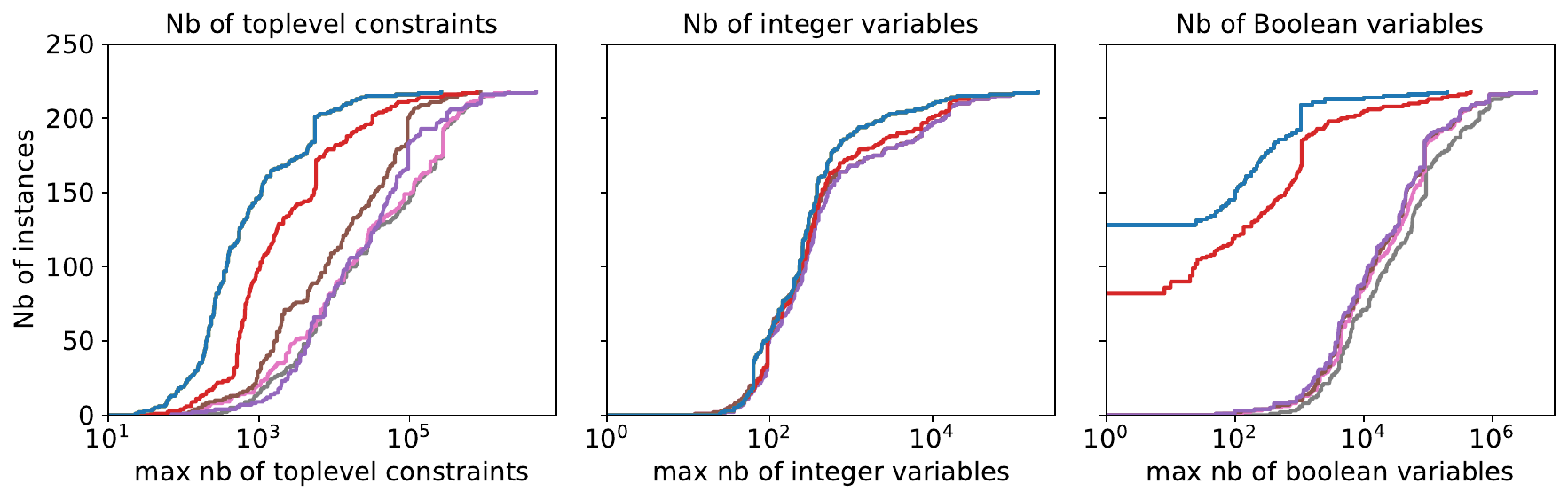}
    \end{minipage}
    \caption{Cumulative distribution plots on the number of constraints (left), number of integer variables (center), and number of Boolean variables (right) after each stage in the transformation waterfall when interfacing to a pseudo-Boolean solver.}
    \label{fig:transform-stats}
\end{figure}

In this first experimental question, we transform each of the instances in the benchmark set using the entire transformation stack as described in \Cref{sec:transformations}.
\Cref{fig:transform-stats} shows the distribution of the number of constraints and variables after each step in the transformation stack.
Note that both safening of partial functions and elimination of negation do not introduce any new constraints or variables into the model.
Indeed, for the XCSP3 instances, all partial functions are at the top-level of the constraint model, and hence do not need to be safened before they are decomposed.
For eliminating negation, we identify only one case where additional constraints can be introduced: namely, applying DeMorgan's law of rewriting negated disjunctions.
For the XCSP3 instances we considered, this rule was never applied, and hence no new constraints or variables were introduced by the transformation.

As expected, decomposing global constraints introduced a large number of additional constraints and variables into the model.
While in general we try to avoid introducing functionally defined variables as discussed in \Cref{sec:decompose}, for some global constraints and functions we are required to use auxiliary variables.

During flattening, we introduce an auxiliary variable for each nested expression in the model.
Hence, we notice a significant increase in both the number of variables and the number of constraints in the model in \Cref{fig:transform-stats}.
Clearly, a large number of these constraints are reification constraints for when Boolean expressions were used as subexpressions (e.g., as part of a decomposition).
This is also confirmed by the resulting number of constraints after applying our detection algorithm for categorical variables (see \Cref{sec:linearize:categorical}).
Indeed, after encoding all categorical variables, all constraints of the form $\textgrammar{Lit} \reifies \textgrammar{IntVar} \textgrammar{Comparison} \textgrammar{Int}$ are eliminated, and we instead post a single constraint that channels the integer variable to a Boolean encoding.
Interestingly, as shown in \Cref{tab:transform-stats}, the number of Boolean variables increased by only a minor amount after constructing the Boolean encoding.
This means that almost all values in the integer variable's domain are already used in a variable-equals-value subexpression elsewhere in the model.
Still, the average number of constraints is reduced by 63\% for the models we tested.
Note that the number of integer variables also slightly decreases after this transformation.
Indeed, when interfacing to a purely Boolean solver, we do not post the channeling constraint between the Boolean encoding and the integer variable at this stage of the transformation stack.
Hence, if an integer variable only occurs in reified comparisons with a constant, it will be removed from the model after encoding the categorical variables with a direct encoding.

During linearization of the constraints, no new integer variables are introduced, as all linear constraints can easily be normalized by altering the coefficients of linear terms and/or by swapping comparisons.
All constraints and Boolean variables introduced by this transformation are due to introducing big-M constraints for linearizing half-reified linear inequalities and encoding disequality constraints, as discussed in \Cref{sec:linearize}.

In the final step of the transformation waterfall, we encode all integer variables with an appropriate Boolean encoding.
Interestingly, both the number of constraints and Boolean variables increase only moderately compared to the linear formulation of the constraint model.

Overall, we conclude that reformulating a high-level constraint model to a low-level formulation such as Pseudo-Boolean constraints can significantly increase the number of constraints and auxiliary variables, and that detecting categorical variables as part of the linearization transformation is essential to avoid an explosion in the number of constraints and Boolean variables during the final encoding step.

\begin{table}[t]
\centering
\caption{Average increase in number of top-level constraints and variables compared to the input and previous transformation stage. Top-level conjunctions are split into their individual arguments and counted as individual constraints.}
\begin{tabular}{l||cc|cc|cc}
 & \multicolumn{2}{c|}{Number of constraints} & \multicolumn{2}{c|}{Number of integer variables} & \multicolumn{2}{c}{Number of Boolean variables} \\
Compared to & Input & Previous & Input & Previous & Input & Previous \\
\midrule
Input & $\times 1.0$ & $\times 1.0$ & $\times 1.0$ & $\times 1.0$ & $\times 1.0$ & $\times 1.0$ \\
No partial functions & $\times 1.0$ & $\times 1.0$ & $\times 1.0$ & $\times 1.0$ & $\times 1.0$ & $\times 1.0$ \\
Eliminate negation & $\times 1.0$ & $\times 1.0$ & $\times 1.0$ & $\times 1.0$ & $\times 1.0$ & $\times 1.0$ \\
Decompose global & $\times 3.9$ & $\times 3.9$ & $\times 1.4$ & $\times 1.4$ & $\times 4.9$ & $\times 4.9$ \\
Flatten & $\times 27.5$ & $\times 7.0$ & $\times 1.8$ & $\times 1.3$ & $\times 53.7$ & $\times 11.1$ \\
Detect categorical & $\times 10.3$ & $\times 0.4$ & $\times 1.8$ & $\times 1.0$ & $\times 54.5$ & $\times 1.0$ \\
Linearize & $\times 31.1$ & $\times 3.0$ & $\times 1.8$ & $\times 1.0$ & $\times 57.2$ & $\times 1.1$ \\
Integer to Boolean & $\times 32.8$ & $\times 1.1$ & $\times 0.0$ & $\times 0.0$ & $\times 85.2$ & $\times 1.5$ \\
\bottomrule
\end{tabular}
\label{tab:transform-stats}
\end{table}

\subsection{EQ2: impact of transformation optimizations on solve time for ILP, PB and SAT Solvers}
\begin{figure}[t]
    \centering
    \begin{minipage}[c]{0.19\linewidth}
    \includegraphics[width=\linewidth]{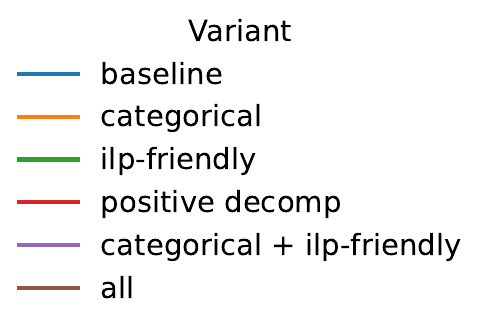}    
    \end{minipage}%
    \begin{minipage}[c]{0.81\linewidth}
    \includegraphics[width=\linewidth]{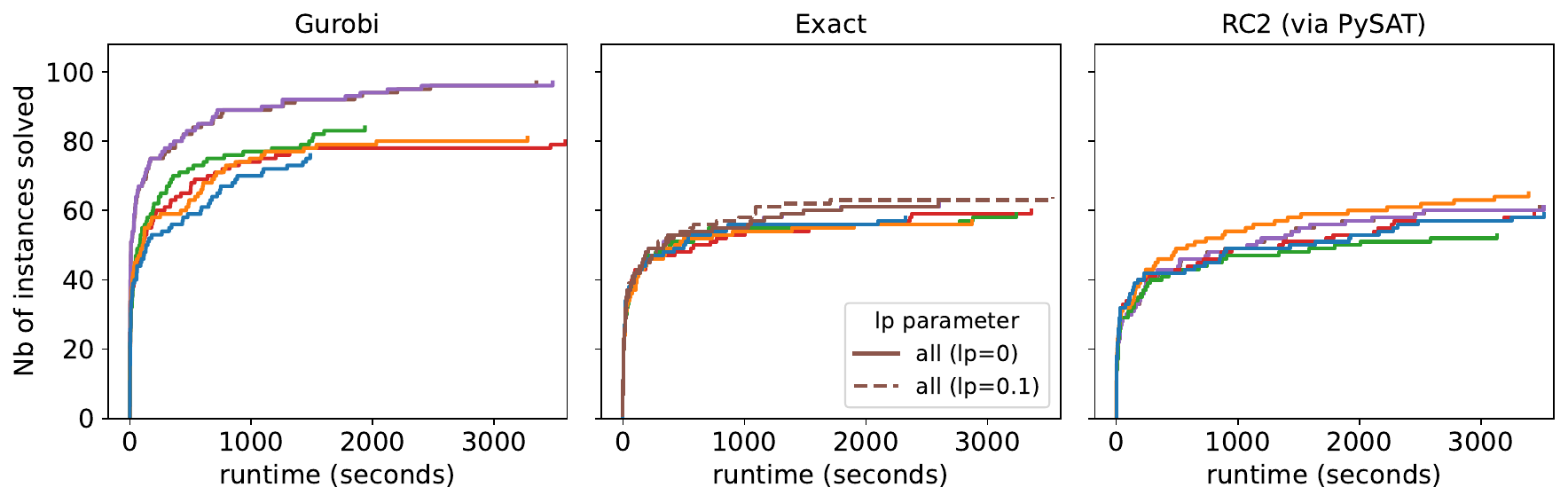}
    \end{minipage}
    \caption{Runtime to proven optimal solution when enabling different optimizations in the transformation pipeline.}
    \label{fig:ablation-results}
\end{figure}

In this second experimental question, we evaluate the impact of the transformation optimizations described in \Cref{sec:transformations} on the solving time of lower-level solvers.
\Cref{fig:ablation-results} shows the runtime until proven optimal for the constraint optimization models in our benchmark. The time limit was set to 1h.
We now discuss the effects of different optimizations for each solver.

\paragraph{Integer Linear Programming solvers}
We first focus on the results for ILP solvers on the left-side of \Cref{fig:ablation-results}.
Here we generally notice a positive effect for any of the optimizations we implemented.
Indeed, compared to the baseline (blue), using specialized positive decompositions improves the solving performance, as well as using specialized ILP-friendly decompositions that are known to have a good linear relaxation.
Interestingly, this is even the case without automatically encoding the categorical variables using their direct encoding (\Cref{sec:linearize:categorical}).
Note that Gurobi supports implication constraints in their API and does the big-M encoding of these constraints internally.
Therefore, it is likely that Gurobi encodes some of the categorical variables itself when the solver detects it.
Still, combining our own direct encoding for categorical variables, in combination with the ILP-friendly decompositions result in the most instances solved to optimality overall.
Adding the specialized positive decompositions on top have little to no impact, as most ILP-friendly decompositions work in both positive and nested contexts.

\paragraph{Pseudo-Boolean solvers}

Compared to the results for ILP solvers, the optimizations implemented in our pipeline have a smaller impact on the result for PB solvers.
Indeed, even the combination of all optimizations perform only marginally better compared to the baseline.
Note that Exact also supports to use an LP-solver during the solving process, which is used to solve the linear relaxation of the constraint model to provide (strong) bounds.
When we allow the solver to spend 10\% of the solving time in the LP-solver (lp = 0.1), the solving time is slightly improved when using all optimizations in our pipeline.

\paragraph{Max-SAT solvers}

Finally, we discuss the results for Max-SAT solvers as shown on the right-hand side of \Cref{fig:ablation-results}.
A surprising result here is the negative impact of using ILP-friendly decompositions for global constraints.
Indeed, especially in the absence of eagerly encoding categorical variables using a direct encoding, the solver performance degrades compared to the baseline CP-style decompositions.
This is likely due to the large domains that can be induced by some ILP-friendly decompositions.
For example, our pipeline rewrites constraint $\element{[10,20,30]}{i}  \geq r$ to $\set{10\directenc{i}{1} + 20\directenc{i}{2} + 30\directenc{i}{3} = r', r' \geq r}$, the auxiliary variable $r'$ will have domain \range{0}{60}, and the direct encoding will make a Boolean literal for each of these domain values.
Naturally, while the consistency constraints will exclude all except 10, 20 and 30 from its domain, the Boolean literals are still posted to the solver.
Further optimizations to the transformation pipeline could include automatically detecting such cases and already shrinking the domain of $r'$ during the reformulation process.

Overall, the current best combination is to use the eager encoding for categorical variables in combination with standard CP-style decompositions.

\section{Conclusion}
We presented a modular end-to-end transformation pipeline for finite-domain constraint models. It can target CP, SMT, ILP, PB, MaxSAT, and SAT solvers by incrementally rewriting high-level Boolean and integer expressions into the subset of the language supported by each solver. The pipeline is organized as a waterfall of reusable transformations: eliminating partial functions, eliminating negation, decomposing unsupported global constraints and functions, flattening nested expressions, linearizing constraints, encoding integer variables as Boolean literals, and translating pseudo-Boolean constraints to CNF as needed.

The main challenges involved correctly preserving the semantics of negated and nested expressions, while avoiding unnecessary creation of auxiliary variables. In particular, partial functions must be safened before later transformations change their Boolean context and decompositions must remain valid when global constraints occur in nested contexts. To keep the reformulated models compact, we delayed the explicit introduction of auxiliary variables where possible, used functional decompositions for global functions and constraints, and applied common subexpression elimination and normalization throughout.

We experimentally evaluated the transformation pipeline on 250 optimization instances from the XCSP3 competition. The results show that constraint models change substantially throughout the transformation stack, especially during decomposition, flattening, and linearization. We also observe that transformation optimizations are important for solving performance: ILP-friendly decompositions, direct encodings for categorical integer variables, and positive-context decompositions can significantly reduce model size and improve solving behavior for ILP, PB, and SAT-based backends.

Future work can extend the input language, for example to floating-point decision variables.
It can also improve individual transformations, such as normalization, half-reification-aware decompositions, and CSE, or investigate different hierarchical transformation architectures altogether.
As the experimental results show, the encoding to (Max)SAT does not necessarily benefit from the same transformations as ILP/PB solvers.
Hence, a more direct path from integer expressions, like PicatSAT~\cite{zhou2015picat} implements, could be of benefit.
Finally, we notice a trend that more and more solvers support limited forms of nested expressions, for example nested affine transformations of integers for OR-Tools CP-SAT~\cite{perron2022ortools}, nested integer expressions for Gurobi~13~\cite{url:gurobi}, and nested Boolean expressions for PySDD ~\cite{darwiche2017pysdd} and Paramita~\cite{alos2026paramita}.
This encourages further reducing the importance and work of the flatten step, which has been a central component of previous CP modeling languages.

\begin{acks}
This research is partly funded by the European Research Council (ERC) under the EU Horizon 2020 research and innovation programme (Grant No 101002802, CHAT-Opt).
Additionally, this research received funding from the framework of H.F.R.I call “4th Call for H.F.R.I.’s
Research Projects to Support Postdoctoral Researchers” (H.F.R.I. Project Number: 28553). 
\end{acks}

\printbibliography

\appendix

\input{appendix}

\FloatBarrier
\end{document}

%% file: auxiliary.tex
\newcommand{\pageplan}[1]{{\gmall{\textit{\color{red!50!black} PP: #1}}}}
\renewcommand{\pageplan}[1]{}

\newcommand{\ignore}[1]{}

\renewcommand{\implies}{\Rightarrow}
\newcommand{\equals}{$=$}

\newcommand{\cons}[2]{\textsc{#1}(#2)}
\newcommand{\term}[1]{\mbox{\textsc{#1}}\xspace}
\newcommand{\expr}{\varepsilon}

\newcommand{\alldiff}[1]{\cons{AllDifferent}{#1}}

\newcommand{\gcc}[1]{\cons{GlobalCardinality}{#1}}
\newcommand{\conscount}[1]{\cons{Count}{#1}}
\newcommand{\nvalue}[1]{\cons{NValue}{#1}}
\newcommand{\atmostnvalue}[1]{\cons{AtMostNValue}{#1}}

\newcommand{\argmin}[1]{\cons{ArgMin}{#1}}
\newcommand{\consdiv}[2]{\cons{Div}{#1,#2}}
\newcommand{\consmod}[2]{\cons{Modulo}{#1,#2}}

\newcommand{\consmin}[1]{\cons{Min}{#1}}
\newcommand{\element}[2]{\cons{Element}{#1,#2}}
\renewcommand{\element}[2]{#1[#2]}
\newcommand{\regular}[1]{\cons{Regular}{#1}}
\newcommand{\constable}[1]{\cons{Table}{#1}}
\newcommand{\negtable}[1]{\cons{NegativeTable}{#1}}
\newcommand{\mddcons}[1]{\cons{MDD}{#1}}

\newcommand{\tcount}{\term{Count}}
\newcommand{\tdiv}{\term{Division}}
\newcommand{\tcircuit}{\term{Circuit}}
\newcommand{\targmin}{\term{ArgMin}}
\newcommand{\talldiff}{\term{AllDifferent}}

\newcommand{\telement}{\term{Element}}
\newcommand{\tcumulative}{\term{Cumulative}}
\newcommand{\tgcc}{\term{GlobalCardinalityCount}}
\newcommand{\tnvalue}{\term{NValue}}
\newcommand{\tmax}{\term{Max}}
\newcommand{\tmin}{\term{Min}}
\newcommand{\tabs}{\term{Abs}}
\newcommand{\txor}{\term{Xor}}

\newcommand{\tequals}{$=$\xspace}

\newcommand{\tregular}{\term{Regular}}
\newcommand{\ttable}{\term{Table}}
\newcommand{\tmdd}{\term{MDD}}
\newcommand{\tsum}{\term{Sum}}

\newcommand{\textgrammar}[1]{{~\mathit{\langle #1 \rangle}~}\xspace}

\newcommand{\gcomparison}{\textgrammar{Comparison}}

\renewcommand{\implies}{\Rightarrow}

\newcommand{\ttrue}{\emph{true}\xspace}
\newcommand{\tfalse}{\emph{false}\xspace}
\newcommand{\reifies}{\Leftrightarrow}

\usepackage{amsmath}
\usepackage{xspace}
\newcommand\mm[1]{\ensuremath{#1}\xspace}
\newcommand\call[1]{\mm{\textsc{#1}}}

\newcommand{\lb}[1]{\lfloor #1 \rfloor}
\renewcommand{\lb}[1]{\mathit{lb}(#1)}
\newcommand{\ub}[1]{\lceil #1 \rceil}
\renewcommand{\ub}[1]{\mathit{ub}(#1)}

\newcommand{\set}[1]{\ensuremath{\{#1\}}}

\newcommand{\variables}{\mm{\mathcal{X}}}

\newcommand{\variablex}{\mm{X}}
\newcommand{\variabley}{\mm{Y}}
\newcommand{\xvar}{\variablex}
\newcommand{\yvar}{\variabley}
\newcommand{\domains}{\mathcal{D}}

\newcommand{\domainof}[1]{D_{#1}}

\newcommand{\range}[2]{[#1..#2]}

\newcommand{\constraints}{\mathcal{C}}

\newcommand{\scope}[1]{\mathit{scope}(#1)}

\newcommand{\assignmentto}[1]{\alpha_{|#1}}

\newcommand{\sols}[1]{\mathit{sols}(#1)}
\newcommand{\solsto}[2]{\mathit{sols}_{#1}(#2)}

\newcommand{\assign}[2]{#1 \mapsto #2}

\newcommand{\choco}{Choco\xspace}

\newcommand{\pindakaas}{Pindakaas\xspace}
\newcommand{\pypblib}{PyPBlib\xspace}

\newcommand\valuecons[2]{\mm{\mathcal{V}(#1,#2)}}
\newcommand\auxvars{\mm{\mathcal{A}}}
\newcommand\definingcons[2]{\mm{\mathcal{T}(#1,#2)}}
\newcommand\definingtotal[2]{\mm{\mathcal{T}_F(#1,#2)}}

\newcommand{\edgevarval}[3]{e_{#1 \xrightarrow{#3} #2}}
\renewcommand{\edgevarval}[3]{e(#1 \xrightarrow{#3} #2)}
\renewcommand{\edgevarval}[3]{\llbracket #1 \mathrel{\xrightarrow{\!#3\!}} #2 \rrbracket}
\newcommand\glob[1]{\mm{G(#1)}}
\newcommand\globx{\glob{\variables}}

\newcommand{\supported}{\mm{\mathit{Supp}}}
\newcommand{\supportednested}{\mm{\mathit{SuppNested}}}

\newcommand{\directencB}[2]{\mm{B_{#1 = #2}}}
\newcommand{\directenc}[2]{\mm{\llbracket#1 = #2\rrbracket}}
\newcommand{\orderenc}[2]{\mm{\llbracket#1 \geq #2\rrbracket}}
\newcommand{\logenc}[2]{\mm{\llbracket\mathit{bit}(#1,#2)\rrbracket}}

\newcommand{\csemap}{\mm{\mathit{csemap}}}

\newcommand{\bv}{\mm{b}}

%% file: images/waterfall.tex
\usepackage{tikz}
\usetikzlibrary{arrows.meta, positioning, fit}

\tikzset{
  >={Latex[length=2mm]},
  base/.style = {
    rectangle,
    draw=black,
    minimum width=2.6cm,
    minimum height=0.8cm,
    align=center,
    font=\small,
  },
  transf/.style = {
    base,
    fill=black!5
  },
  start/.style = {
    font=\small\itshape
  },
  arrow/.style = {
    ->,
    thick
  },
  group/.style args={#1}{
    draw=black,
    dashed,
    rounded corners,
    label={[anchor=south west, xshift=2pt, yshift=5pt]south west:#1},
    inner sep=5pt
  },
}

\newcommand{\transformations}{
\begin{figure}[b]
\centering
\begin{tikzpicture}[
    node distance=0.5cm and -0.9cm
  ]

  \node (model)      [start] {model};

  \node (safen)      [transf, below right=of model]      {No partial functions};
  \node (negation)   [transf, below right=of safen]      {Eliminate negation};
  \node (decompose)  [transf, below right=of negation]   {Decompose global};
  \node (flatten)    [transf, below right=of decompose]  {Flatten};
  \node (linearize)  [transf, below right=of flatten]    {Linearize};
  \node (int2bool)   [transf, below right=of linearize]  {Integer to Bool};
  \node (cnfnode)    [transf, below right=of int2bool]   {To CNF};

  \draw[arrow] (model.east) to[out=0, in=90] (safen.north);
  \draw[arrow] (safen.east) to[out=0, in=90] (negation.north);
  \draw[arrow] (negation.east) to[out=0, in=90] (decompose.north);
  \draw[arrow] (decompose.east) to[out=0, in=90] (flatten.north);
  \draw[arrow] (flatten.east) to[out=0, in=90] (linearize.north);
  \draw[arrow] (linearize.east) to[out=0, in=90] (int2bool.north);
  \draw[arrow] (int2bool.east) to[out=0, in=90] (cnfnode.north);

  \draw[arrow, dotted] ([xshift=-1cm]safen.south) to[out=-90, in=180] (decompose.west);

  \node[group={SMT-solvers}, fit=(model)(decompose)] {};
  \node[group={CP-solvers},  inner sep=7pt, xshift=-1pt, yshift=1pt, fit=(model)(flatten)] {};
  \node[group={ILP-solvers}, inner sep=8pt, xshift=-3pt, yshift=3pt, fit=(model)(linearize)] {};
  \node[group={PB-solvers},  inner sep=9pt, xshift=-5pt, yshift=5pt, fit=(model)(int2bool)] {};
  \node[group={SAT-solvers}, inner sep=11pt, xshift=-7pt, yshift=7pt, fit=(model)(cnfnode)] {};

\end{tikzpicture}
\caption{Transformation pipeline.}
\label{fig:transformations}
\end{figure}
}

%% file: grammars/core-alt.tex
\begin{Grammar}[b]
\caption{The constraint modeling language.}
\label{gra:cpmpy}
\begin{grammar}
    <Model> ::= "[" <BoolExpr> ("," <BoolExpr>)* "], minimize=" <IntExpr>

    <BoolExpr> ::= $ $
            \alt <Bool>
            \alt <BoolVar>
            \alt $\lnot$ <BoolExpr>
            \alt <BoolExpr> $\land$ | $\lor$ | $\Rightarrow$ | $\Leftrightarrow$ <BoolExpr>
            \alt <IntExpr> $\leq$ | \textless{} | $\geq$ | \textgreater{} | \equals{} | $\neq$ <IntExpr> 
            \alt <GlobalConstraint> "[" <Arg> ("," <Arg>)* "]"

    <IntExpr> ::= $ $
            \alt <Int>
            \alt <IntVar> \label{intvar}
            \alt <BoolExpr>
            \alt "-" <IntExpr>
            \alt <IntExpr> "+" | "-" <IntExpr>
            \alt <Int>\texttimes<IntExpr>
            \alt <GlobalFunction> "[" <Arg> ("," <Arg>)* "]"
    
    <BoolVar> ::= <String>   

    <IntVar> ::= <String> <Int> <Int>
    
    <Arg> ::= <BoolExpr> | <IntExpr> | "[" <Arg> ("," <Arg>)* "]"
\end{grammar}
\end{Grammar}

%% file: images/partials.tex
\tikzset{
  constraint tree/.style={
    level distance=7mm,
    sibling distance=1.2cm,
    every node/.style={
      shape=rectangle,
      draw,
      align=center,
      minimum height=5mm
    },
    nbc node/.style={
      draw=blue,
      text=blue
    },
    execute at end picture={
      \path[use as bounding box] (0,-24mm) rectangle (0,0mm);
    }
  }
}

\newcommand{\topleveldiv}{%
\begin{tikzpicture}[
    baseline=(current bounding box.north),
    constraint tree
]\node[nbc node] (root) {\small =}
    child { node {\small \term{Div}}
        [sibling distance=0.6cm]
        child { node {\small $x$}}
        child { node {\small $y$}}
    }
    child { node {\small $z$}};

\end{tikzpicture}%
}

\newcommand{\nestedsumdiv}{%
\begin{tikzpicture}[
    baseline=(current bounding box.north),
    constraint tree
]\node[nbc node] (root) {\small =}
    child { node {\small +}
        [sibling distance=0.8cm]
        child { node {\small \term{Div}}
            [sibling distance=0.6cm]
            child { node {\small $x$}}
            child { node {\small $y$}}
        }
        child { node {\small $d$}}
    }
    child { node {\small $z$}};

\end{tikzpicture}%
}

\newcommand{\nesteddiv}{%
\begin{tikzpicture}[
    baseline=(current bounding box.north),
    constraint tree
]\node (root) {\small $\vee$}
    child { node[nbc node] {\small =}
        [sibling distance=0.8cm]
        child { node {\small \term{Div}}
            [sibling distance=0.6cm]
            child { node {\small $x$}}
            child { node {\small $y$}}
        }
        child { node {\small $z$}}
    }
    child { node {\small $p$}};

\end{tikzpicture}%
}

\newcommand{\negateddiv}{%
\begin{tikzpicture}[
    baseline=(current bounding box.north),
    constraint tree
]\node (root) {\small $\neg$}
    child { node[nbc node] {\small =}
        [sibling distance=0.8cm]
        child { node {\small \term{Div}}
            [sibling distance=0.6cm]
            child { node {\small $x$}}
            child { node {\small $y$}}
        }
        child { node {\small $z$}}
    };

\end{tikzpicture}%
}

%% file: grammars/negation.tex
\begin{Grammar}[h]
\caption{Boolean expressions after eliminating negation.}
\label{gram:negation}
\begin{grammar}
    <BoolExpr> ::= $ $
            \alt <Bool>
            \alt <BoolVar>
            \alt {\color{blue} $\neg$<BoolVar>}
            \alt <BoolExpr> $\land$ | $\lor$ | $\Rightarrow$ | $\Leftrightarrow$ <BoolExpr>
            \alt <IntExpr> $\leq$ | \textless{} | $\geq$ | \textgreater{} | \equals{} | $\neq$ <IntExpr> 
            \alt <GlobalConstraint> "[" <Arg> ("," <Arg>)* "]"
            \alt {\color{blue}$\neg$<GlobalConstraint> "[" <Arg> ("," <Arg>)* "]"}
\end{grammar}
\end{Grammar}

%% file: algorithms/flatten.tex
\begin{algorithm}[h]
\caption{\call{Flatten($C$, $\csemap$)}}
\label{algo:flatten}
\begin{algorithmic}[1]
    \State \textbf{Input:} set of constraints $C$, CSE map $\csemap$
    \State \textbf{Output:} set of flat constraints
    \State $F \gets \emptyset$;\; $\todolist \gets \emptyset$
    \For{$c \in C$}
        \State $c \gets \call{SimplifyOperators}(c)$
        \If{$c$ is a $\textgrammar{FnfBoolExpr}$}
            \State $F \gets F \cup \set{c}$\; \textbf{continue}
        \EndIf
        \If{$c$ is $\expr_1 \circ \expr_2$ with $\circ \in \set{\reifies, \implies}$}
            \State $\expr_1', d_1 \gets \getormakevar{\expr_1}{\csemap}$
            \State $\expr_2', d_2 \gets \normalizedboolexpr{\expr_2}{\csemap}$
            \State $F \gets F \cup \set{\expr_1' \circ \expr_2'}$
            \State $\todolist \gets \todolist \cup d_1 \cup d_2$
        \ElsIf{$c$ is $expr_1 \circ \expr_2$ with $\circ \in  \set{=, \neq, <, >, \leq, \geq}$}
            \State $\expr_1', d_1 \gets \getormakevar{\expr_1}{\csemap}$
            \State $\expr_2', d_2 \gets \normalizednumexpr{\expr_2}{\csemap}$
            \State $F \gets F \cup \set{\expr_1' \circ \expr_2'}$
            \State $\todolist \gets \todolist \cup d_1 \cup d_2$
        \ElsIf{$c$ is $\neg\,\expr$} \Comment{$\expr$ can only be a global constraint after eliminating negation}
            \State $\expr', d \gets \normalizedboolexpr{\expr'}{\csemap}$
            \State $F \gets F \cup \set{\neg\,\expr}$
            \State $\todolist \gets \todolist \cup d$
        \Else \Comment{other $\textgrammar{BoolExpr}$}
            \State $c', d \gets \normalizedboolexpr{c, \csemap}$
            \State $F \gets F \cup \set{c'}$
            \State $\todolist \gets \todolist \cup d$
        \EndIf
    \EndFor
    \If{$\todolist \neq \emptyset$} 
         $F \gets F \cup \call{Flatten}(\todolist, \csemap)$
    \EndIf
    \State \Return $F$
\end{algorithmic}
\end{algorithm}

%% file: algorithms/getormakevar.tex
\begin{algorithm}[h]
\caption{\call{GetOrMakeVar($\expr, \csemap$)}}
\label{algo:getormakevar}
\begin{algorithmic}[1]

    \If {$\expr$ is a variable}
         \Return $\expr, \emptyset$
    \EndIf

    \State $\expr \gets \textsc{Normalize}(\expr)$
    
    \If{$\expr \in \csemap$}
        \State \Return $\mathit{\csemap.get(\expr)}, \emptyset$
    \EndIf
    
    \If{$\textsc{IsBoolExpr}(\expr)$}
        \State $v \gets \mathit{boolvar()}$
        \State $d \coloneq v \reifies \expr$
    \Else
        \State $v \gets \mathit{intvar}(\lb{\expr}, \ub{\expr})$
        \State $d \coloneq v = \expr$
    \EndIf

    \State $\mathit{\csemap.put(\expr, v)}$

    \State \Return $v, d$
\end{algorithmic}
\end{algorithm}

%% file: grammars/flat-alt.tex
\begin{Grammar}[t!]
\caption{Flat normal form}
\label{grammar:flat2}
\begin{grammar}
    
    <Model> ::= "[" <FnfBoolExpr> ("," <FnfBoolExpr>)* "], minimize=" <Int> + <LinExpr>

    <Lit> ::= <BoolVar> | $\lnot$ <BoolVar>

    <FnfBoolExpr> ::= $ $
        \alt <PrimitiveExpr>
        \alt <Lit> $\implies$ <PrimitiveExpr>
        \alt <Lit> $\reifies$ <PrimitiveExpr>

    <PrimitiveExpr> ::= $ $
        \alt False
        \alt <Lit> ($\lor$ <Lit>)*  % n-ary clause
        \alt <LinExpr> <Comparison> <Int>
        \alt <IntVar> <Comparison> <GlobalFunction> "[" <FnfArg> ("," <FnfArg>)* "]"
        \alt <GlobalConstraint> "[" <FnfArg> ("," <FnfArg>)* "]"
        \alt $\neg$ <GlobalConstraint> "[" <FnfArg> ("," <FnfArg>)* "]"

    <LinExpr> ::= $ $
        \alt <Var> ("+" <Var>)*
        \alt <Int>"\texttimes"<Var> ("+" <Int>"\texttimes"<Var>)*

    <Var> ::= <BoolVar> | <IntVar>
    
    <Comparison> ::= $\leq$ | $\textless{}$ | $\geq$ | $\textgreater{}$ | \equals{} | $\neq$
    
    <FnfArg> ::= <Int> | <Var> | "[" <FnfArg> ("," <FnfArg>)* "]"

\end{grammar}
\end{Grammar}

%% file: grammars/only-numexpr-equality.tex
\begin{Grammar}[t!]
\caption{Primitive expressions after normalizing global functions}
\label{grammar:equality}
\begin{grammar}
    <PrimitiveExpr> :: = $ $
        \alt False
        \alt <Lit> ($\lor$ <Lit>)*  % n-ary clause
        \alt <LinExpr> <Comparison> <Int>
        \alt {\color{blue}<IntVar> \equals{} <GlobalFunction> "[" <FnfArg> ("," <FnfArg>)* "]"}
        \alt <GlobalConstraint> "[" <FnfArg> ("," <FnfArg>)* "]"
        \alt $\neg$ <GlobalConstraint> "[" <FnfArg> ("," <FnfArg>)* "]"
\end{grammar}
\end{Grammar}

%% file: grammars/only-implies.tex
\begin{Grammar}[t!]
\caption{Flat Boolean expressions after normalizing reification.}
\label{grammar:reification}
\begin{grammar}

    <FnfBoolExpr> ::= $ $
        \alt <PrimitiveExpr>
        \alt <Lit> $\implies$ <PrimitiveExpr>
        \alt {\color{blue}<Lit> $\reifies$ <IntVar> <Comparison> <Int>}

\end{grammar}
\end{Grammar}

%% file: grammars/linear.tex
\begin{Grammar}[b]
\caption{Linear normal form}
\label{grammar:linear}
\begin{grammar}
    <Model> ::= "[" <LinCons> ("," <LinCons>)* "], minimize=" <Int> + <LinExpr>

    <LinCons> ::= $ $
            \alt False
            \alt <LinExpr> <Comparison> <Int>

    <LinExpr> ::= $ $
        \alt <Var> ("+" <Var>)*
        \alt <Int>"\texttimes"<Var> ("+" <Int>"\texttimes"<Var>)*
    
    <Var> ::= <BoolVar> | <IntVar>
    
    <Comparison> ::=  \equals{} | $\leq$ | $\geq$
\end{grammar}
\end{Grammar}

%% file: appendix.tex
\section{Pseudocode for Normalization of Arguments During Flattening}
\label{appendix:normalize}
\input{algorithms/normalized-boolexpr}
\input{algorithms/normalized-numexpr}

%% file: algorithms/normalized-boolexpr.tex
\begin{algorithm}[H]
\caption{$\normalizedboolexpr{\expr}{\csemap}$}
\begin{algorithmic}[1]
\If{$\expr$ is a variable or constant}
    \State \Return ($\expr, \emptyset$)
\ElsIf{$\expr$ is a $\textgrammar{Comparison}$ $\expr_l \circ \expr_r$}
    \State $\expr_l', d_l \gets \getormakevar{\expr_l}{\csemap}$
    % \If{$\expr_l$ is Boolean}
    %     \State $\expr_r', d_r \gets \getormakevar{\expr_r}{\csemap}$
    % \Else
    \State $\expr_r', d_r \gets \normalizednumexpr{\expr_r}{\csemap}$
    % \EndIf
    \If {$\expr_r$ is not a $\textgrammar{GlobalFunction}$}
        \State $\expr_l', \expr_r' \gets \call{CanonicalizedLinear}(\expr_l', \expr_r')$
    \EndIf
    \State \Return $(\expr_l' \circ \expr_r',\; d_l \cup d_r)$
\ElsIf{$\expr$ is $\expr_1 \implies \expr_2$}
    \State $\expr_1', d_1 \gets \getormakevar{\expr_1}{\csemap}$
    \State $\expr_2', d_2 \gets \getormakevar{\expr_2}{\csemap}$
    \State \Return $(\neg\,\expr_1' \lor \expr_2',\; d_1 \cup d_2)$
\ElsIf{$\expr$ is $\expr_1 \reifies \expr_2$}
    \State $\expr_1', d_1 \gets \getormakevar{\expr_1}{\csemap}$
    \State $\expr_2', d_2 \gets \getormakevar{\expr_2}{\csemap}$
    \State \Return $(1\times\expr_1' + -1 \times \expr_2' = 0,\; d_1 \cup d_2)$
\ElsIf{$\expr$ is $\neg\,\expr_0$} \Comment{Negation of global constraint}
    \State $\expr_0', d \gets \normalizedboolexpr{\expr_0}{\csemap}$
    \State \Return $(\neg\,\expr_0',\; d)$
\EndIf
\Statex {$\triangleright \vee$, $\reifies$ and global constraints}
\For{each argument $a_i$ of $\expr$}
    \State $v_i, d_i \gets \getormakevar{a_i}{\csemap}$
\EndFor
\State \Return $(\expr[v_i/a_i],\; \bigcup_i d_i)$
\end{algorithmic}
\end{algorithm}

%% file: algorithms/normalized-numexpr.tex
\begin{algorithm}[h]
\caption{\call{NormalizeNumExpr}($\expr$, $\csemap$)}
\begin{algorithmic}[1]
\If{$\expr$ is a variable or constant}
    \State \Return $(\expr, \emptyset)$
\ElsIf{$\expr$ is a Boolean expression}
    \State \Return $\getormakevar{\expr}{\csemap}$
    % \Comment{convert Boolean to integer via reification}
% \ElsIf{$\expr = w \cdot \expr_0$ with constant $w$}
%     \State \Return \Call{NormalizeNumExpr}{$\textgrammar{wsum}(w, \expr_0)$, $\csemap$}
\ElsIf{$\expr$ is $- \expr_0$}
    \State $\expr_0', d_0 \gets \getormakevar{\expr_0}{\csemap}$
    \State \Return ($-1 \times \expr_0', d_0$)
\ElsIf{$\expr$ is $\expr_1 - \expr_2$}
    \State $\expr_1', d_1 \gets \getormakevar{\expr_1}{\csemap}$
    \State $\expr_2', d_2 \gets \getormakevar{\expr_2}{\csemap}$
    \State \Return ($1 \times \expr_1' + -1 \times \expr_2', d_1 \cup d_2$)
\ElsIf{$\expr$ is $\sum_i w_i \times \expr_i$}
    \For {each argument $\expr_i$}
        \State $\expr_i', d_i \gets \getormakevar{\expr_i}{\csemap}$
    \EndFor
    \State \Return $(\sum w_i \times \expr_i', \bigcup_i d_i)$
\EndIf
\Statex $\triangleright +$ and global functions
\For{each argument $a_i$ of $\expr$}
    \State $v_i, d_i \gets \getormakevar{a_i}{\csemap}$
\EndFor
\State \Return $(\expr[v_i/a_i],\; \bigcup_i d_i)$
\end{algorithmic}
\end{algorithm}

%% file: bibliography.bib
@inproceedings{vanhentenryck2014views,
	author = {Pascal {Van Hentenryck} and
                  Laurent D. Michel},
	title = {Domain Views for Constraint Programming},
	booktitle = {{CP}},
	series = {Lecture Notes in Computer Science},
	pages = {705--720},
	publisher = {Springer},
	year = {2014}
}

@article{miller1960integer,
	author = {C. E. Miller and
                  Albert W. Tucker and
                  R. A. Zemlin},
	title = {Integer Programming Formulation of Traveling Salesman Problems},
	journal = {J. {ACM}},
	volume = {7},
	number = {4},
	pages = {326--329},
	year = {1960}
}

@inproceedings{liffiton2012minicard,
	author = {Mark H. Liffiton and
                  Jordyn C. Maglalang},
	title = {A Cardinality Solver: More Expressive Constraints for Free - (Poster
                  Presentation)},
	booktitle = {{SAT}},
	series = {Lecture Notes in Computer Science},
	volume = {7317},
	pages = {485--486},
	publisher = {Springer},
	year = {2012}
}

@inproceedings{marquessilva2007towards,
	author = {Jo{\~{a}}o Marques{-}Silva and
                  In{\^{e}}s Lynce},
	title = {Towards Robust {CNF} Encodings of Cardinality Constraints},
	booktitle = {{CP}},
	series = {Lecture Notes in Computer Science},
	volume = {4741},
	pages = {483--497},
	publisher = {Springer},
	year = {2007}
}

@incollection{prestwich2021cnfencodings,
	author = {Steven D. Prestwich},
	title = {{CNF} Encodings},
	booktitle = {Handbook of Satisfiability},
	series = {Frontiers in Artificial Intelligence and Applications},
	volume = {336},
	pages = {75--100},
	publisher = {{IOS} Press},
	year = {2021}
}

@inproceedings{bierlee2025revisiting,
	author = {Hendrik Bierlee and
                  Jip J. Dekker and
                  Peter J. Stuckey},
	title = {Revisiting Pseudo-Boolean Encodings from an Integer Perspective},
	booktitle = {{CPAIOR} {(1)}},
	series = {Lecture Notes in Computer Science},
	volume = {15762},
	pages = {113--133},
	publisher = {Springer},
	year = {2025}
}

@inproceedings{aavani2011translating,
	author = {Amir Aavani},
	title = {Translating Pseudo-Boolean Constraints into {CNF}},
	booktitle = {{SAT}},
	series = {Lecture Notes in Computer Science},
	volume = {6695},
	pages = {357--359},
	publisher = {Springer},
	year = {2011}
}

@inproceedings{walsh2000sat,
	author = {Toby Walsh},
	title = {{SAT} v {CSP}},
	booktitle = {{CP}},
	series = {Lecture Notes in Computer Science},
	volume = {1894},
	pages = {441--456},
	publisher = {Springer},
	year = {2000}
}

@inproceedings{ernst1997automatic,
	author = {Michael D. Ernst and
                  Todd D. Millstein and
                  Daniel S. Weld},
	title = {Automatic SAT-Compilation of Planning Problems},
	booktitle = {{IJCAI}},
	pages = {1169--1177},
	publisher = {Morgan Kaufmann},
	year = {1997}
}

@article{tamura2009compiling,
	author = {Naoyuki Tamura and
                  Akiko Taga and
                  Satoshi Kitagawa and
                  Mutsunori Banbara},
	title = {Compiling Finite Linear {CSP} into {SAT}},
	journal = {Constraints An Int. J.},
	volume = {14},
	number = {2},
	pages = {254--272},
	year = {2009}
}

@article{schwerin1997binpacking,
	title = {The Bin-Packing Problem: A Problem Generator and Some Numerical Experiments with FFD Packing and MTP},
	author = {Petra Schwerin and Gerhard W{\"a}scher},
	journal = {International transactions in operational research},
	volume = {4},
	number = {5-6},
	pages = {377--389},
	year = {1997},
	publisher = {Elsevier}
}

@article{hoeve2001alldifferent,
	author = {Willem-Jan {van Hoeve}},
	title = {The {AllDifferent} Constraint: {A} Survey},
	journal = {CoRR},
	volume = {cs.PL/0105015},
	year = {2001}
}

@article{rendl2008eliminating,
	title = {Eliminating Common Subexpressions during Flattening},
	author = {Andrea Rendl and Ian P Gent and Ian Miguel},
	journal = {URL: http://www-circa. mcs. st-and. ac. uk/Preprints/ERCIMCSE08. pdf},
	year = {2008}
}

@article{jefferson2010implementing,
	author = {Christopher Jefferson and
                  Neil C. A. Moore and
                  Peter Nightingale and
                  Karen E. Petrie},
	title = {Implementing Logical Connectives in Constraint Programming},
	journal = {Artif. Intell.},
	volume = {174},
	number = {16-17},
	pages = {1407--1429},
	year = {2010}
}

@article{beldiceanu2013reification,
	author = {Nicolas Beldiceanu and
                  Mats Carlsson and
                  Pierre Flener and
                  Justin Pearson},
	title = {On the Reification of Global Constraints},
	journal = {Constraints An Int. J.},
	volume = {18},
	number = {1},
	pages = {1--6},
	year = {2013}
}

@article{prudhomme2016choco,
	title = {Choco Solver Documentation},
	author = {Charles Prud’homme and Jean-Guillaume Fages and Xavier Lorca},
	journal = {TASC, INRIA Rennes, LINA CNRS UMR},
	volume = {6241},
	pages = {102--106},
	year = {2016}
}

@article{akgun2022conjure,
	title = {Conjure: Automatic Generation of Constraint Models from Problem Specifications},
	author = {{\"O}zg{\"u}r Akg{\"u}n and Alan M Frisch and Ian P Gent and Christopher Jefferson and Ian Miguel and Peter Nightingale},
	journal = {Artificial Intelligence},
	volume = {310},
	pages = {103751},
	year = {2022},
	publisher = {Elsevier}
}

@book{biere2021handbook,
	editor = {Armin Biere and
                  Marijn Heule and
                  Hans {van Maaren} and
                  Toby Walsh},
	title = {Handbook of Satisfiability - Second Edition},
	series = {Frontiers in Artificial Intelligence and Applications},
	volume = {336},
	publisher = {{IOS} Press},
	year = {2021}
}

@book{zhou2015picat,
	author = {Neng{-}Fa Zhou and
                  H{\aa}kan Kjellerstrand and
                  Jonathan Fruhman},
	title = {Constraint Solving and Planning with Picat},
	series = {Springer Briefs in Intelligent Systems},
	publisher = {Springer},
	year = {2015}
}

@inproceedings{ignatiev2018pysat,
	author = {Alexey Ignatiev and
                  Ant{\'{o}}nio Morgado and
                  Jo{\~{a}}o Marques{-}Silva},
	title = {PySAT: {A} Python Toolkit for Prototyping with {SAT} Oracles},
	booktitle = {{SAT}},
	series = {Lecture Notes in Computer Science},
	volume = {10929},
	pages = {428--437},
	publisher = {Springer},
	year = {2018}
}

@book{rossi2006handbook,
	editor = {Francesca Rossi and
                  Peter {van Beek} and
                  Toby Walsh},
	title = {Handbook of Constraint Programming},
	series = {Foundations of Artificial Intelligence},
	volume = {2},
	publisher = {Elsevier},
	year = {2006},
	url = {https://www.sciencedirect.com/science/bookseries/15746526/2},
	isbn = {978-0-444-52726-4},
	bibsource = {dblp computer science bibliography, https://dblp.org}
}

@incollection{smith2006modelling,
	author = {Barbara M. Smith},
	editor = {Francesca Rossi and
                  Peter {van Beek} and
                  Toby Walsh},
	title = {Modelling},
	booktitle = {Handbook of Constraint Programming},
	series = {Foundations of Artificial Intelligence},
	volume = {2},
	pages = {377--406},
	publisher = {Elsevier},
	year = {2006},
	url = {https://doi.org/10.1016/S1574-6526(06)80015-5},
	doi = {10.1016/S1574-6526(06)80015-5},
	bibsource = {dblp computer science bibliography, https://dblp.org}
}

@incollection{hoeve2006global,
	author = {Willem-Jan {van Hoeve} and Irit Katriel},
	booktitle = {Handbook of Constraint Programming},
	title = {Global Constraints},
	doi = {10.1016/S1574-6526(06)80010-6},
	editor = {Francesca Rossi and Peter {van Beek} and Toby Walsh},
	pages = {169--208},
	publisher = {Elsevier},
	series = {Foundations of Artificial Intelligence},
	volume = {2},
	bibsource = {dblp computer science bibliography, https://dblp.org},
	year = {2006}
}

@inproceedings{barrett2010smt,
	title = {The {SMT-LIB} Standard: Version 2.0},
	author = {Barrett, Clark and Stump, Aaron and Tinelli, Cesare and others},
	booktitle = {Proceedings of the 8th International Workshop on Satisfiability Modulo Theories (Edinburgh, {UK})},
	volume = {13},
	pages = {14},
	year = {2010}
}

@inproceedings{vanhoeve2006open,
	author = {Willem-Jan {van Hoeve} and
                  Jean{-}Charles R{\'{e}}gin},
	title = {Open Constraints in a Closed World},
	booktitle = {{CPAIOR}},
	series = {Lecture Notes in Computer Science},
	volume = {3990},
	pages = {244--257},
	publisher = {Springer},
	year = {2006}
}

@inproceedings{vanhentenryck2003be,
	title = {To Be or Not to Be... a Global Constraint},
	author = {Christian Bessiere and Pascal {Van Hentenryck}},
	booktitle = {International Conference on Principles and Practice of Constraint Programming},
	pages = {789--794},
	year = {2003},
	organization = {Springer}
}

@inproceedings{nethercote2007minizinc,
	title = {MiniZinc: Towards a Standard CP Modelling Language},
	author = {Nicholas Nethercote and Peter J Stuckey and Ralph Becket and Sebastian Brand and Gregory J Duck and Guido Tack},
	booktitle = {International Conference on Principles and Practice of Constraint Programming},
	pages = {529--543},
	year = {2007},
	organization = {Springer}
}

@inproceedings{feydy2009why,
	author = {Thibaut Feydy and Peter J. Stuckey and Mark Wallace},
	booktitle = {Principles and Practice of Constraint Programming - {CP} 2009, 15th International Conference, {CP} 2009, Lisbon, Portugal, September 20-24, 2009, Proceedings},
	title = {Why Cumulative Decomposition Is Not as Bad as It Sounds},
	doi = {10.1007/978-3-642-04244-7_58},
	editor = {Ian P. Gent},
	pages = {746--761},
	publisher = {Springer},
	series = {Lecture Notes in Computer Science},
	volume = {5732},
	bibsource = {dblp computer science bibliography, https://dblp.org},
	year = {2009}
}

@inproceedings{frisch2009proper,
	title = {The Proper Treatment of Undefinedness in Constraint Languages},
	author = {Alan M Frisch and Peter J Stuckey},
	booktitle = {International Conference on Principles and Practice of Constraint Programming},
	pages = {367--382},
	year = {2009},
	organization = {Springer}
}

@inproceedings{bessiere2010nvalue,
	title = {Decomposition of the NValue Constraint},
	author = {Christian Bessiere and George Katsirelos and Nina Narodytska and Claude-Guy Quimper and Toby Walsh},
	booktitle = {International Conference on Principles and Practice of Constraint Programming},
	pages = {114--128},
	year = {2010},
	organization = {Springer}
}

@inproceedings{feydy2011half,
	author = {Thibaut Feydy and
                  Zoltan Somogyi and
                  Peter J. Stuckey},
	title = {Half Reification and Flattening},
	booktitle = {{CP}},
	series = {Lecture Notes in Computer Science},
	volume = {6876},
	pages = {286--301},
	publisher = {Springer},
	year = {2011}
}

@incollection{li2021maxsat,
	author = {Chu Min Li and
                  Felip Many{\`{a}}},
	title = {MaxSAT, Hard and Soft Constraints},
	booktitle = {Handbook of Satisfiability},
	series = {Frontiers in Artificial Intelligence and Applications},
	volume = {336},
	pages = {903--927},
	publisher = {{IOS} Press},
	year = {2021}
}

@inproceedings{bleukx2025modeling,
	author = {Ignace Bleukx and
                  Ryma Boumazouza and
                  Tias Guns and
                  Nadine Laage and
                  Guillaume Pov{\'{e}}da},
	title = {Modeling and Explaining an Industrial Workforce Allocation and Scheduling
                  Problem},
	booktitle = {{CP}},
	series = {LIPIcs},
	volume = {340},
	pages = {6:1--6:24},
	publisher = {Schloss Dagstuhl - Leibniz-Zentrum f{\"{u}}r Informatik},
	year = {2025}
}

@article{audemard2025xcsp3,
	author = {Gilles Audemard and
                  Christophe Lecoutre and
                  Emmanuel Lonca},
	title = {Proceedings of the 2025 {XCSP3} Competition},
	journal = {CoRR},
	volume = {abs/2511.06918},
	year = {2025}
}

@article{darwiche2017pysdd,
	author = {Adnan Darwiche and
                  Pierre Marquis and
                  Dan Suciu and
                  Stefan Szeider},
	title = {Recent Trends in Knowledge Compilation (Dagstuhl Seminar 17381)},
	journal = {Dagstuhl Reports},
	volume = {7},
	number = {9},
	pages = {62--85},
	year = {2017}
}

@article{audemard2024xcsp3,
	author = {Gilles Audemard and
                  Christophe Lecoutre and
                  Emmanuel Lonca},
	title = {Proceedings of the 2024 {XCSP3} Competition},
	journal = {CoRR},
	volume = {abs/2412.00117},
	year = {2024}
}

@article{vielma2015mipformulations,
	author = {Juan Pablo Vielma},
	title = {Mixed Integer Linear Programming Formulation Techniques},
	journal = {{SIAM} Rev.},
	volume = {57},
	number = {1},
	pages = {3--57},
	year = {2015}
}

@inproceedings{vanbeek1999cplan,
	author = {Peter {van Beek} and
                  Xinguang Chen},
	title = {CPlan: {A} Constraint Programming Approach to Planning},
	booktitle = {{AAAI/IAAI}},
	pages = {585--590},
	publisher = {{AAAI} Press / The {MIT} Press},
	year = {1999}
}

@book{dantzig1963linear,
	author = {Dantzig, George B.},
	address = {Princeton (N.J.)},
	language = {eng},
	lccn = {59-50559},
	publisher = {Princeton university press},
	title = {Linear Programming and Extensions},
	year = {1963}
}

@inproceedings{achterberg2008scip,
	Title = {Constraint Integer Programming: A New Approach to Integrate CP and MIP},
	Author = {Achterberg, Tobias and Berthold, Timo and Koch, Thorsten and Wolter, Kati},
	Booktitle = {Integration of AI and OR Techniques in Constraint Programming for Combinatorial Optimization Problems},
	Year = {2008},
	Address = {Berlin, Heidelberg},
	Editor = {Perron, Laurent and Trick, Michael A.},
	Pages = {6--20},
	Publisher = {Springer Berlin Heidelberg},
	ISBN = {978-3-540-68155-7},
	doi = {https://doi.org/10.1007/978-3-540-68155-7_4}
}

@inproceedings{stuckey2022enumerated,
	author = {Peter J. Stuckey and
                  Guido Tack},
	title = {Enumerated Types and Type Extensions for MiniZinc},
	booktitle = {{CPAIOR}},
	series = {Lecture Notes in Computer Science},
	volume = {13292},
	pages = {374--389},
	publisher = {Springer},
	year = {2022}
}

@book{ahuja1993network,
	author = {Ravindra K. Ahuja and
                  Thomas L. Magnanti and
                  James B. Orlin},
	title = {Network Flows - Theory, Algorithms and Applications},
	publisher = {Prentice Hall},
	year = {1993}
}

@inproceedings{contaldo2020minizinc2omt,
	author = {Francesco Contaldo and
                  Patrick Trentin and
                  Roberto Sebastiani},
	title = {From MiniZinc to Optimization Modulo Theories, and Back},
	booktitle = {{CPAIOR}},
	series = {Lecture Notes in Computer Science},
	volume = {12296},
	pages = {148--166},
	publisher = {Springer},
	year = {2020}
}

@article{debacker2000vrp,
	author = {Bruno {De Backer} and
                  Vincent Furnon and
                  Paul Shaw and
                  Philip Kilby and
                  Patrick Prosser},
	title = {Solving Vehicle Routing Problems Using Constraint Programming and
                  Metaheuristics},
	journal = {J. Heuristics},
	volume = {6},
	number = {4},
	pages = {501--523},
	year = {2000}
}

@inproceedings{fromherz2001scheduling,
	author = {Markus P. J. Fromherz},
	title = {Constraint-Based Scheduling},
	booktitle = {{ACC}},
	pages = {3231--3244},
	publisher = {{IEEE}},
	year = {2001}
}

@article{clautiaux2008orthogonal,
	author = {Fran{\c{c}}ois Clautiaux and
                  Antoine Jouglet and
                  Jacques Carlier and
                  Aziz Moukrim},
	title = {A New Constraint Programming Approach for the Orthogonal Packing Problem},
	journal = {Comput. Oper. Res.},
	volume = {35},
	number = {3},
	pages = {944--959},
	year = {2008}
}

@incollection{barrett2021smt,
	author = {Clark W. Barrett and
                  Roberto Sebastiani and
                  Sanjit A. Seshia and
                  Cesare Tinelli},
	title = {Satisfiability Modulo Theories},
	booktitle = {Handbook of Satisfiability},
	series = {Frontiers in Artificial Intelligence and Applications},
	volume = {336},
	pages = {1267--1329},
	publisher = {{IOS} Press},
	year = {2021}
}

@incollection{roussel2021pseudoboolean,
	author = {Olivier Roussel and
                  Vasco Manquinho},
	title = {Pseudo-Boolean and Cardinality Constraints},
	booktitle = {Handbook of Satisfiability},
	series = {Frontiers in Artificial Intelligence and Applications},
	volume = {336},
	pages = {1087--1129},
	publisher = {{IOS} Press},
	year = {2021}
}

@article{hart2011pyomo,
	author = {William E. Hart and
                  Jean{-}Paul Watson and
                  David L. Woodruff},
	title = {Pyomo: Modeling and Solving Mathematical Programs in Python},
	journal = {Math. Program. Comput.},
	volume = {3},
	number = {3},
	pages = {219--260},
	year = {2011}
}

@article{nightingale2022savilerow,
	author = {Peter Nightingale},
	title = {Savile Row Manual},
	journal = {CoRR},
	volume = {abs/2201.03472},
	year = {2022}
}

@inproceedings{belov2016improved,
	author = {Gleb Belov and
                  Peter J. Stuckey and
                  Guido Tack and
                  Mark Wallace},
	title = {Improved Linearization of Constraint Programming Models},
	booktitle = {{CP}},
	series = {Lecture Notes in Computer Science},
	volume = {9892},
	pages = {49--65},
	publisher = {Springer},
	year = {2016}
}

@inproceedings{stuckey2013minizincwithfunctions,
	author = {Peter J. Stuckey and
                  Guido Tack},
	title = {MiniZinc with Functions},
	booktitle = {{CPAIOR}},
	series = {Lecture Notes in Computer Science},
	volume = {7874},
	pages = {268--283},
	publisher = {Springer},
	year = {2013}
}

@article{demorgan1864,
	author = {{De Morgan}, Augustus},
	address = {Cambridge},
	journal = {Transactions of the Cambridge Philosophical Society},
	language = {eng},
	number = {1},
	pages = {173-},
	publisher = {Cambridge University Press},
	title = {"On the Syllogism", No. III., "and on Logic in General"},
	volume = {10},
	year = {1864}
}

@inproceedings{bierlee2022coupling,
	author = {Hendrik Bierlee and
                  Graeme Gange and
                  Guido Tack and
                  Jip J. Dekker and
                  Peter J. Stuckey},
	title = {Coupling Different Integer Encodings for {SAT}},
	booktitle = {{CPAIOR}},
	series = {Lecture Notes in Computer Science},
	pages = {44--63},
	publisher = {Springer},
	year = {2022}
}

@inproceedings{soh2015hybrid,
	author = {Takehide Soh and
                  Mutsunori Banbara and
                  Naoyuki Tamura},
	title = {A Hybrid Encoding of {CSP} to {SAT} Integrating Order and Log Encodings},
	booktitle = {{ICTAI}},
	pages = {421--428},
	publisher = {{IEEE} Computer Society},
	year = {2015}
}

@incollection{pblib,
	year = {2015},
	isbn = {978-3-319-24317-7},
	booktitle = {Theory and Applications of Satisfiability Testing -- SAT 2015},
	volume = {9340},
	series = {Lecture Notes in Computer Science},
	editor = {Heule, Marijn and Weaver, Sean},
	doi = {10.1007/978-3-319-24318-4_2},
	title = {{PBLib} -- A Library for Encoding Pseudo-Boolean Constraints into {CNF}},
	publisher = {Springer International Publishing},
	author = {Philipp, Tobias and Steinke, Peter},
	pages = {9-16}
}

@software{pindakaas,
	author = {Bierlee, Hendrik and Dekker, Jip J.},
	license = {MPL-2.0},
	title = {{Pindakaas}},
	url = {https://doi.org/10.5281/zenodo.10851855},
	doi = {10.5281/zenodo.10851855}
}

@inproceedings{alos2026paramita,
	author = {Josep Al{\`o}s and Carlos Ans{\'o}tegui and Juan Luis Esteban and Eduard Torres},
	title = {Paramita: An Extensible Framework for SATisfiability Solving},
	booktitle = {ModRef 2026},
	year = {2026}
}

@misc{globalconstraintcatalog,
	title = {Global Constraint Catalog, (Revision a)},
	author = {Nicolas Beldiceanu and Mats Carlsson and Jean-Xavier Rampon},
	year = {2012},
	publisher = {Swedish Institute of Computer Science}
}

@misc{exact,
	title = {Exact Solver},
	author = {Jo Devriendt},
	url = {https://gitlab.com/JoD/exact},
	year = {2023}
}

@misc{url:gurobi,
	author = {Gurobi Optimization},
	title = {{Gurobi Optimizer Reference Manual}},
	year = {2026},
	url = {https://www.gurobi.com}
}

@phdthesis{fages2012reifying,
	title = {Reifying Global Constraints},
	author = {Fran{\c{c}}ois Fages and Sylvain Soliman},
	year = {2012},
	school = {INRIA}
}

@software{perron2022ortools,
	title = {OR-Tools},
	version = { v9.6 },
	author = {Laurent Perron and Vincent Furnon},
	organization = {Google},
	url = {https://developers.google.com/optimization/},
	year = {2022},
	month = {11},
	day = {25}
}

@techreport{beldiceanu2012reification,
	author = {Nicolas Beldiceanu and 
              Mats Carlsson and 
              Pierre Flener and 
              Justin Pearson},
	title = {On the Reification of Global Constraints},
	year = {2012},
	type = {Technical report},
	number = {T2012:02}
}


%% file: toappear.bib
@article{bleukx2026halfreified,
	author = {Ignace Bleukx and 
                  H{\'e}l{\`e}ne Verhaeghe and 
                  Dimos Tsouros and 
                  Tias Guns},
	title = {Efficient Reformulations of Half-Reified Global Constraints Using Auxiliary Variables},
	journal = {J. Artif. Intell. Res.},
	year = {2026},
	note = {to appear}
}

@inproceedings{sergeys2026unified,
	author = {Thomas Sergeys and Ignace Bleukx and Tias Guns},
	title = {Unified Programmatic Access to CO Benchmarks, to Connect Constraint Solving Communities},
	booktitle = {SAT},
	series = {LIPIcs},
	publisher = {Schloss Dagstuhl - Leibniz-Zentrum fur Informatik},
	year = {2026},
	note = {to appear}
}

@inproceedings{foschini2026hmlv,
	author = {Marco Foschini and
                  Emilio Gamba and
                  Luca Kletzander and
                  Tias Guns},
	title = {From {CP} Modeling to Preference Elicitation in {HMLV} Assembly Problems},
	booktitle = {{CP}},
	series = {LIPIcs},
	publisher = {Schloss Dagstuhl - Leibniz-Zentrum f{\"{u}}r Informatik},
	year = {2026},
	note = {to appear}
}
